\documentclass[11pt]{article}

\usepackage[final]{acl}

\usepackage{times}
\usepackage{latexsym}
\usepackage[T1]{fontenc}
\usepackage[utf8]{inputenc}
\usepackage{microtype}
\usepackage{subcaption}
\usepackage{inconsolata}
\usepackage{graphicx}
\usepackage{booktabs}
\usepackage{amssymb}
\usepackage{multirow}
\usepackage{multicol}
\usepackage{adjustbox}
\usepackage{pifont}
\usepackage{amsmath}
\usepackage{algorithmicx,algpseudocode,algorithm}
\usepackage{cleveref}
\usepackage{stfloats}
\usepackage{array}
\usepackage{xcolor}
\usepackage{colortbl}
\usepackage[most]{tcolorbox}

\usepackage[most]{tcolorbox}
\tcbset{
  promptbox/.style={
    colback=gray!5,
    colframe=gray!60,
    sharp corners,
    boxrule=0.7pt,
    left=8pt,
    right=8pt,
    top=8pt,
    bottom=8pt,
  }
}
\newcommand{\xmark}{\ding{55}}%
\title{OmniFusion: Simultaneous Multilingual and Multimodal Translations via Modular Fusion}

\author{Sai Koneru$^{1,3}$\thanks{Work done at Karlsruhe Institute of Technology.},
  Matthias Huck$^{2}$, \textnormal{and}
  Jan Niehues$^{1}$ \\
  $^{1}$ Karlsruhe Institute of Technology \\
  $^{2}$ SAP SE, Dietmar-Hopp-Allee 16, 69190 Walldorf, Germany \\
  $^{3}$ AppTek, Aachen, Germany \\
  \texttt{\{sai.koneru, jan.niehues\}@kit.edu} \\
  \texttt{\{matthias.huck\}@sap.com}}

\begin{document}
\maketitle
\begin{abstract}
Open-source text-only translation LLMs have rapidly improved multilingual coverage and translation quality, but their unimodal design limits their use in multimodal translation. In speech translation, they are commonly used in cascaded ASR–MT pipelines, which introduce additional latency and prevent the translation model from directly using visual context. Conversely, pretrained multimodal foundation models provide strong audio-visual perception but generally lag behind specialized translation LLMs in multilingual generation quality. We introduce OmniFusion, a modular architecture that connects a frozen multimodal foundation model to a translation-specialized LLM through a lightweight gated fusion module. OmniFusion supports speech-to-text, speech-and-image-to-text, and text-and-image-to-text translation within a single framework. Experiments with Qwen2.5-Omni-7B and Seed-X-PPO-7B show that selected early and middle MMFM layers provide more transferable representations than late layers. On MCIF, OmniFusion improves the latency–quality trade-off over a matched cascaded SimulST pipeline when evaluated with Local Agreement decoding, while remaining competitive in offline ST and reducing severe XCOMET-XL errors. On CoMMuTE, it achieves competitive visually grounded caption translation performance.
\footnote{Code is available at \url{https://github.com/saikoneru/OmniFusion}}
\end{abstract}

\section{Introduction}

In recent years, NLP has shifted from unimodal foundation models
\citep{Touvron2023-rq, Grattafiori2024-sv} to multimodal foundation models
(MMFMs) \citep{Abdin2024-ez, Goel2025-mv, Xu2025-fv}. While unimodal systems
work well in some settings, many tasks benefit from additional cross-modal cues.

Translation is one such task. Speech Translation (ST) converts spoken input into
a target language \citep{ahmad-etal-2024-findings, agostinelli2025findings},
while caption translation often requires image context for disambiguation.
However, using visual context such as images or presentation slides in ST remains
underexplored \citep{Liu2024-um, Gaido2024-qu, Sinhamahapatra2025-cf}.

\begin{figure*}[!h]
    \centering
    \includegraphics[width=\linewidth]{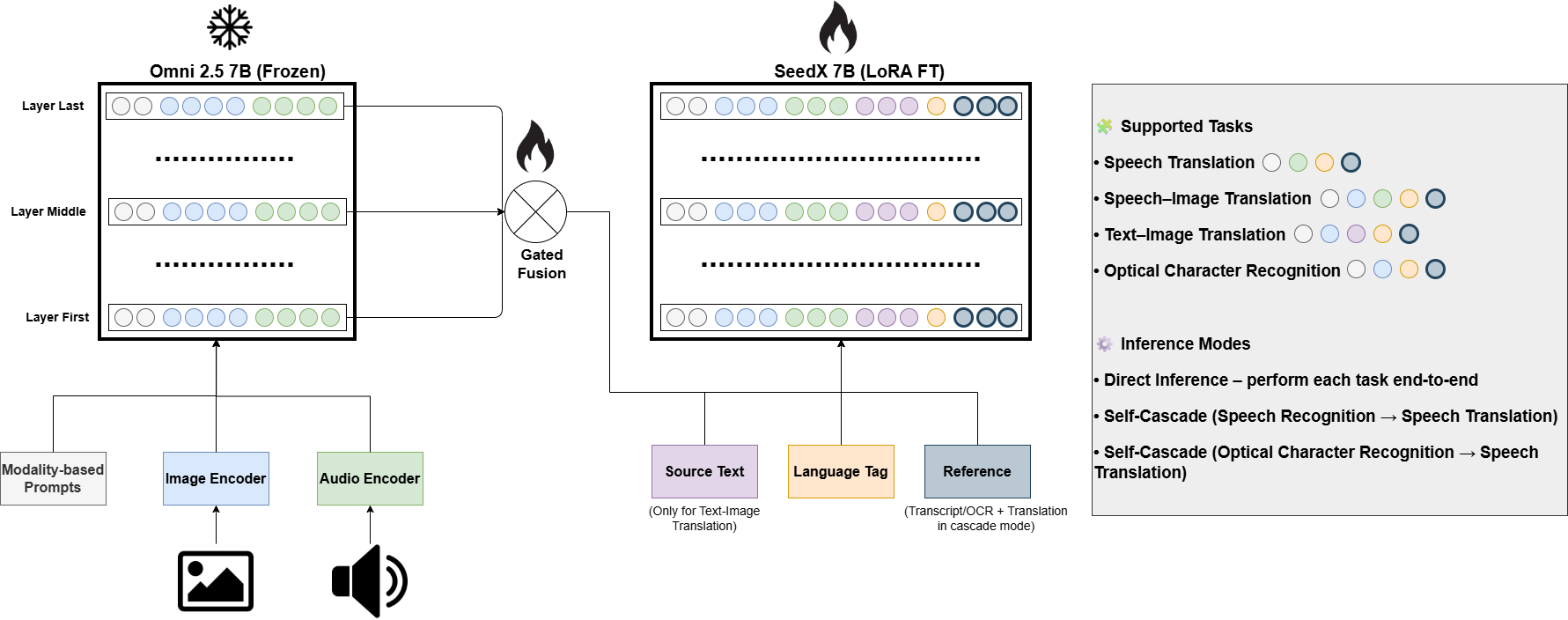}
    \caption{Architecture of OmniFusion with supported tasks and inference modes.}
    \label{fig:placeholder}
\end{figure*}

Although MMFMs process multimodal inputs, their multilingual translation quality
is generally weaker than specialized translation LLMs
\citep{Alves2024-ls, Cheng2025-id}. In ST, cascaded ASR--MT pipelines
\citep{ahmad-etal-2024-findings, agostinelli2025findings, Koneru2025-gz}
remain strong, but they cannot directly exploit multimodal context and add
latency, especially in simultaneous ST (SimulST)
\citep{agostinelli2025findings, Papi2025-mm}.

These limitations motivate an end-to-end (E2E) multilingual and multimodal
translation system for ST, image-aware ST, and caption translation. Existing
work typically treats speech translation, speech--image translation, and
text--image translation as separate tasks, rather than as a unified multimodal
translation problem. Prior work \citep{Ambilduke2025-kh, Viveiros2025-dc} often
trains models from scratch or uses only pretrained encoders, leaving the internal
cross-modal representations of MMFMs underutilized.

We address this gap with OmniFusion, which uses a frozen MMFM as a multimodal
representation source for a translation-specialized LLM. OmniFusion learns a
lightweight fusion interface and uses multitask training with OCR and ASR
self-cascading to improve multimodal alignment and generation.

Our central hypothesis is that MMFM hidden states contain useful cross-modal
signals that can enable all multimodal translation tasks with limited training.

\vspace{1mm}

\textbf{Our contributions are:}
\textbf{(1)} We propose OmniFusion, a modular architecture connecting a frozen
MMFM to a translation-specialized LLM, enabling speech-to-text,
speech-and-image-to-text, and text-and-image-to-text translation.
\textbf{(2)} We show that MMFM layer choice is critical where different 
layers transfer better for different translation tasks and gated fusion is important.
\textbf{(3)} OmniFusion with Local Agreement decoding \citep{liu2020low}
improves the SimulST latency--quality trade-off over the same cascade
(Figure~\ref{fig:simul_st}), reduces major and critical offline ST errors
(Table~\ref{tab:averaged_xcomet}), and achieves competitive CoMMuTE performance
(Table~\ref{tab:commute_comet}).

\section{Approach}

We propose a modular fusion framework that combines a frozen multimodal foundation model, $\mathcal{M}_{\mathrm{MMFM}}$, for audio and image encoding with a translation-specialized language model, $\mathcal{M}_{\mathrm{NMT}}$, for multilingual generation. A lightweight fusion module selects MMFM hidden representations and projects them into the translation LLM’s embedding space, enabling direct conditioning on audio and visual inputs.

Our main architectural contribution is a \emph{gated fusion module} (\S\ref{sec:gated_fusion}) that integrates MMFM representations into the translation LLM. We motivate this design through layer-probing analysis (\S\ref{sec:mmfm_probe}), which shows that early and middle MMFM layers transfer best for translation. We improve performance through the fusion architecture and multitask training with self-cascading ASR and OCR objectives (\S\ref{sec:training}). The system is trained as an offline E2E translation model and applied to SimulST at inference time using Local Agreement decoding \citep{liu2020low}. Figure~\ref{fig:placeholder} illustrates the system architecture and supported tasks.

\subsection{Architecture}
\label{sec:arch}

Let $\mathcal{M}_{\mathrm{NMT}}$ denote the translation model and 
$\mathcal{M}_{\mathrm{MMFM}}$ denote the multimodal foundation model. 
Both models use a language-model backbone. As a reference, standard text-only NMT trains
$\mathcal{M}_{\mathrm{NMT}}$ on source sentence $X=(x_1,\dots,x_{L_x})$ and
target sentence $Y=(y_1,\dots,y_{L_y})$ with the teacher-forcing
cross-entropy objective:
\begin{equation}
\label{eq:trans_loss}
\begin{aligned}
\mathcal{L}_{\mathrm{Trans}}
&=
-\frac{1}{L_y}
\sum_{t=1}^{L_y}
\log P_{\theta}
\left(
y_t \mid y_{<t}, X
\right),
\end{aligned}
\end{equation}
where $X = (x_1,\dots,x_{L_x})$ and $Y = (y_1,\dots,y_{L_y})$ denote the
source and target token sequences, respectively.

For multimodal translation, image and audio inputs are encoded by
$\mathcal{M}_{\mathrm{MMFM}}$ and projected into hidden states compatible with
its language-model backbone. Our goal is to adapt the translation model to use
these representations as additional context.

\subsubsection{Probing MMFM Representations}
\label{sec:mmfm_probe}

\begin{figure}[!ht]
    \centering
    \includegraphics[width=\linewidth]{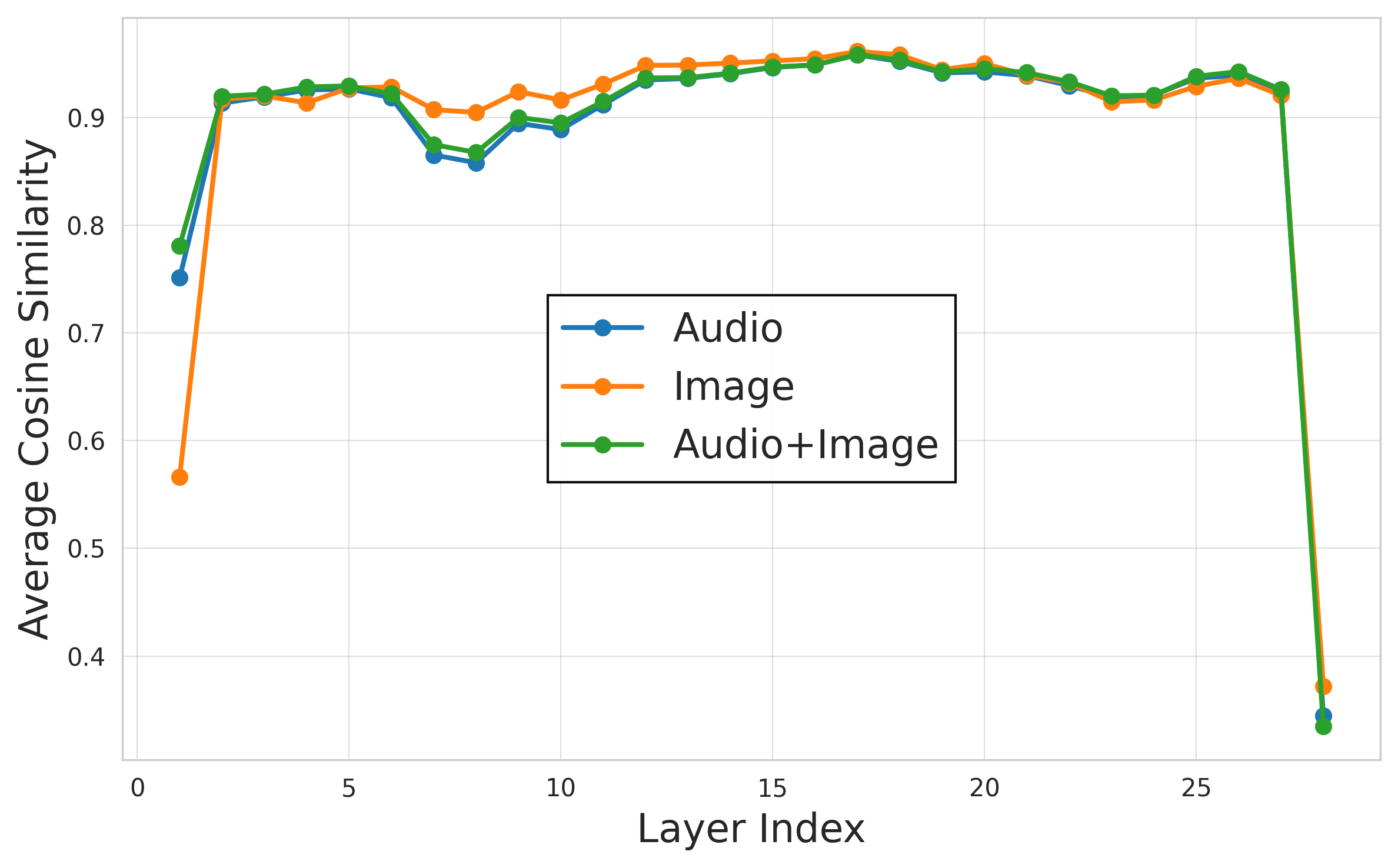}
    \caption{
    Average cosine similarity between hidden representations of adjacent
    layers, computed with Qwen-Omni 2.5 7B on the MCIF test set
    (en $\rightarrow$ de) under different modality conditions.
    }
    \label{fig:layer_analysis}
\end{figure}

Different MMFM layers may expose different information to the translation LLM.
We use adjacent-layer cosine similarity as a diagnostic to identify candidate
layers with distinct representational behavior (\Cref{fig:layer_analysis}).
This diagnostic does not prove what semantic information each layer encodes;
rather, it helps select a small candidate set instead of using all MMFM layers,
which would increase sequence length and may introduce noise. We see the shift occurs drastically at first and last layer indicating three distinct representation regions. Hence, we select the first, middle, and last MMFM layers as candidate
transfer points to cover these regions. Their usefulness is validated through downstream ablations in
Table \ref{tab:fusion_multilinguality} and Appendix Table \ref{tab:layer_analysis}.

\subsection{Gated Layer Fusion}
\label{sec:gated_fusion}

Let $Z = (z_1,\dots,z_{L_m})$ denote the multimodal token sequence produced by
the MMFM input encoder, where $L_m$ is the multimodal sequence length. We denote
by $h_t^{(k)}\in\mathbb{R}^{D_{\mathrm{MMFM}}}$ the hidden state of token
$z_t$ at layer $k$ of $\mathcal{M}_{\mathrm{MMFM}}$, for
$k\in\{\mathrm{first},\mathrm{mid},\mathrm{last}\}$ and $t=1,\dots,L_m$.

Naively feeding all three layer representations as separate context tokens
would triple the multimodal sequence length, treat all layers as equally
informative, and ignore the dimensional mismatch between
$D_{\mathrm{MMFM}}$ and $D_{\mathrm{NMT}}$. We address these with a
\emph{token-wise gated fusion} module.

For each multimodal token $z_t$, we concatenate the three selected layer
representations:
\begin{equation}
\label{eq:layer_concat}
\begin{aligned}
c_t
&=
h_t^{(\mathrm{first})}
\oplus
h_t^{(\mathrm{mid})}
\oplus
h_t^{(\mathrm{last})}, \\
c_t
&\in
\mathbb{R}^{3D_{\mathrm{MMFM}}},
\qquad
t = 1,\dots,L_m ,
\end{aligned}
\end{equation}
where $\oplus$ denotes feature-wise concatenation.

A bias-free linear layer $W_{\mathrm{gate}}\in\mathbb{R}^{3\times 3D_{\mathrm{MMFM}}}$
maps $c_t$ to a three-dimensional logit vector, which is normalized via
softmax to give per-layer importance weights:
\begin{equation}
\label{eq:gate}
\begin{aligned}
\alpha_t
&=
\mathrm{softmax}
\left(
W_{\mathrm{gate}} c_t
\right), \\
\alpha_t
&=
\left[
\alpha_t^{(\mathrm{first})},
\alpha_t^{(\mathrm{mid})},
\alpha_t^{(\mathrm{last})}
\right]
\in
\mathbb{R}^{3}.
\end{aligned}
\end{equation}

The fused MMFM representation is computed as a weighted sum:
\begin{equation}
\label{eq:fused_mmfm}
\begin{aligned}
r_t
&=
\alpha_t^{(\mathrm{first})} h_t^{(\mathrm{first})}
+
\alpha_t^{(\mathrm{mid})} h_t^{(\mathrm{mid})}
+
\alpha_t^{(\mathrm{last})} h_t^{(\mathrm{last})}, \\
r_t
&\in
\mathbb{R}^{D_{\mathrm{MMFM}}},
\qquad
t = 1,\dots,L_m .
\end{aligned}
\end{equation}

Finally, the fused representation is projected into the embedding space of the
translation model using a three-layer MLP for sufficient transformation computation for all modalities with GELU activation:
\begin{equation}
\label{eq:mmfm_projection}
\begin{aligned}
v_t
&=
\mathrm{MLP}(r_t),  \\
v_t
&\in
\mathbb{R}^{D_{\mathrm{NMT}}},
\qquad
t = 1,\dots,L_m .
\end{aligned}
\end{equation}
where $D_{\mathrm{NMT}}$ is the token embedding dimension of
$\mathcal{M}_{\mathrm{NMT}}$.

Let $e_i \in \mathbb{R}^{D_{\mathrm{NMT}}}$ denote the embedding of source token
$x_i$. The projected multimodal embeddings are prepended to the source-token
embeddings (sentence or tag) along the sequence dimension:
\begin{equation}
\label{eq:mm_input}
\begin{aligned}
\widetilde{X}
&=
(v_1,\dots,v_{L_m}, e_1,\dots,e_{L_x}), \\
\widetilde{X}
&\in
\mathbb{R}^{(L_m + L_x) \times D_{\mathrm{NMT}}}.
\end{aligned}
\end{equation}
Thus, the fused multimodal representations are treated as additional input
tokens for the translation model.

The multimodal translation objective is then defined as:
\begin{equation}
\label{eq:mm_trans_loss}
\begin{aligned}
\mathcal{L}_{\mathrm{MM\text{-}Trans}}
&=
-\frac{1}{L_y}
\sum_{t=1}^{L_y}
\log P_{\theta}
\left(
y_t \mid y_{<t}, \widetilde{X}
\right).
\end{aligned}
\end{equation}

\subsection{Training}
\label{sec:training}

We align multimodal features with the translation LLM through multitask training with the objectives below.

\subsubsection{Multi-task Learning}
We train on three tasks. In ST, the model translates source-language audio. In Speech--Image Translation (SIT), it translates audio with an aligned image, such as a presentation slide. In Text--Image Translation (TIT), it translates source text paired with an image. These tasks expose the model to different modality combinations and encourage multimodal alignment.

\subsubsection{Self-Cascading (ASR and OCR)}
In addition to multi-task learning, we adopt self-cascading, which decomposes translation into intermediate steps. For example in English-to-German ST, the model first generates the ASR transcription, then produces the final translation conditioned on it. For visual inputs, OCR text serves as an intermediate step before translation. During training, self-cascading is applied stochastically on the samples with modality-specific prompts for better MMFM representations (\Cref{alg:prompt}). We freeze the MMFM and fine-tune the translation LLM with LoRA \citep{hulora}. We find that adding LoRA to the MMFM degrades performance due to catastrophic forgetting.

We distinguish between training-time and inference-time self-cascading.
When training, self-cascading examples teach the model to generate
intermediate ASR or OCR text before the final translation. At inference
time, we evaluate both direct generation, where the model outputs only
the translation or self-cascade generation.

\section{Experimental Setup}
\subsection{Training Data}

We train OmniFusion on speech- and image-conditioned translation data. An
overview of data sources is provided in Appendix \Cref{tab:data_overview}. For
ST, we use subsets of Europarl-ST \citep{jairsan2020a} and CoVoST
\citep{wang2020covost} across multiple target languages. For image-aware ST, we
use M3AV \citep{chen-etal-2024-m3av}, which contains English scientific talks
paired with slides and Paddle-OCR \citep{cui2025paddleocr} predictions. Since
M3AV has limited language coverage, we augment ASR transcripts with Seed-X-PPO-7B
\citep{Cheng2025-id} translations to create \textbf{M3AV-Aug}, enabling
speech-image translation and self-cascade training. We do not apply
filtering but show that the translation quality is high even when training on this synthetic data. For TIT, we use Multi30k \citep{elliott-etal-2016-multi30k}, which pairs images
with multilingual captions and encourages visual grounding beyond OCR.

\subsection{Models}
Several MMFMs have recently shown strong gains in quality and efficiency \citep{Abdin2024-ez, gemmateam2025gemma3technicalreport}. As our setting requires processing both audio and image modalities with cross-modal reasoning, we use Qwen2.5-Omni 7B\footnote{Qwen/Qwen2.5-Omni-7B} \citep{Xu2025-fv}.
For translation, we adopt Seed-X-PPO-7B\footnote{ByteDance-Seed/Seed-X-PPO-7B} \citep{Cheng2025-id}, which provides broad language coverage and strong translation performance compared to closed-source alternatives. Using the training data described above, we refer to the resulting fine-tuned model as \textbf{OmniFusion}. While we report results with these models, preliminary experiments with alternative model pairings (Appendix Table \ref{tab:layer_analysis}) suggest the approach can extend to other pairings. Training hyperparameters are listed in \Cref{tab:training-hyperparameters}. Validation uses CoVoST dev data, selecting the checkpoint with the lowest loss after 20k steps.

\subsection{Evaluation: Data \& Metrics}

Evaluating multimodal SimulST requires aligned audio–visual translation data. We exclude M3AV from testing due to training overlap and its monolingual nature, and instead assess robustness on unseen domains using the MCIF test set \citep{papi2025mcifmultimodalcrosslingualinstructionfollowing}, which contains ACL talk recordings translated from English into three target languages, and FLEURS \citep{conneau2023fleurs} for multilingual coverage. Latency is reported using Average Lagging \citep{simuleval2020}, and translation quality is measured on these datasets.

For TIT, we target the cases where visual context is crucial for disambiguation, using the CoMMuTE benchmark \citep{futeral-etal-2023-tackling,futeral-etal-2025-towards}. Translation quality is evaluated with XCOMET-XL \citep{guerreiro-etal-2024-xcomet} for ST, enabling detailed error analysis, and with COMET\footnote{Unbabel/wmt22-comet-da} \citep{rei-etal-2022-comet} for TIT.

\section{Results}
\label{sec:results}

\begin{figure*}[!ht]
    \centering
    \resizebox{2\columnwidth}{!}{
    \begin{subfigure}[b]{0.45\textwidth}
        \centering
        \includegraphics[width=\textwidth]{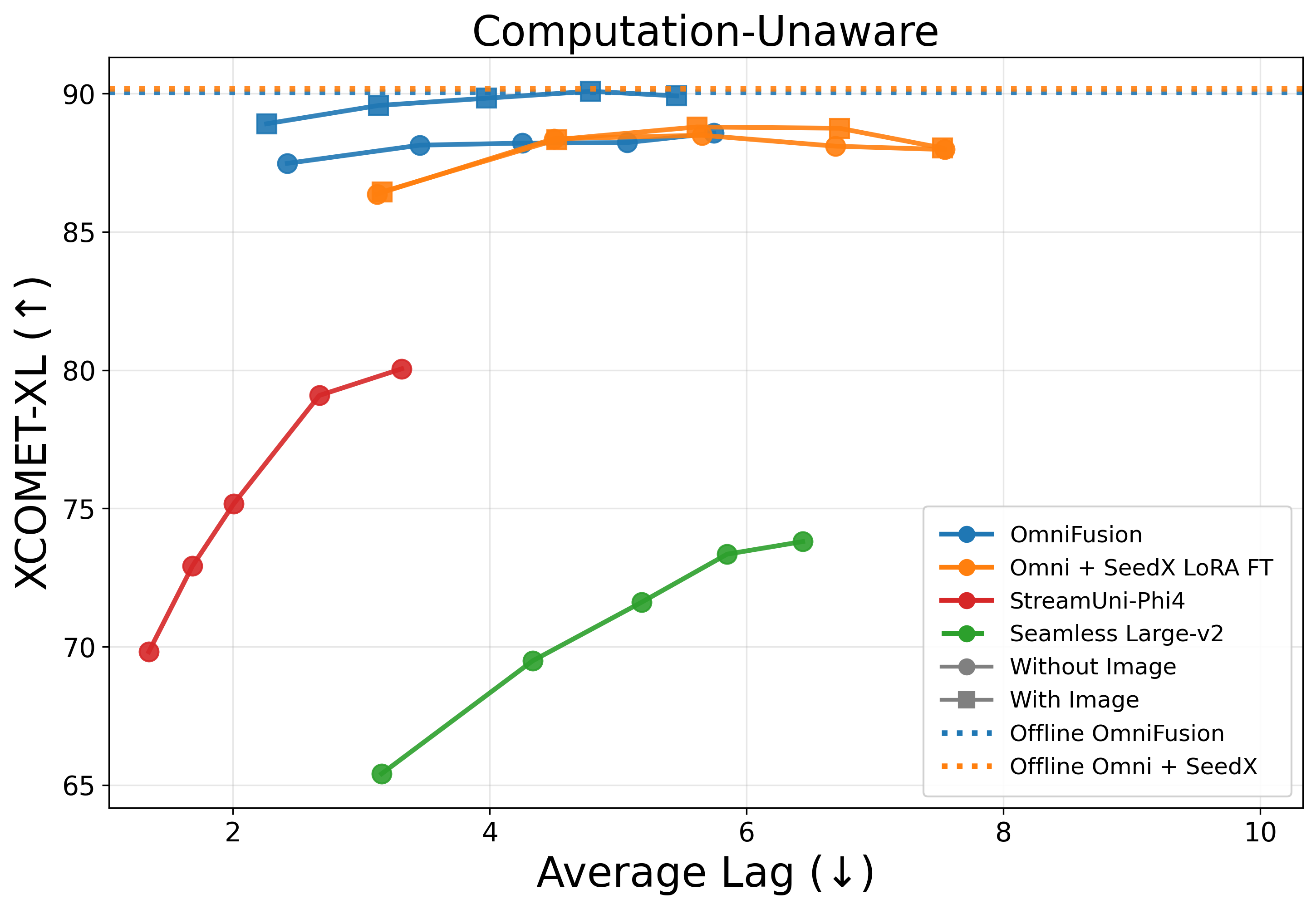}
        \caption{en $\rightarrow$ de}
        \label{fig:simul_1000_image_ende_false}
    \end{subfigure}
    \hfill
    \begin{subfigure}[b]{0.45\textwidth}
        \centering
        \includegraphics[width=\textwidth]{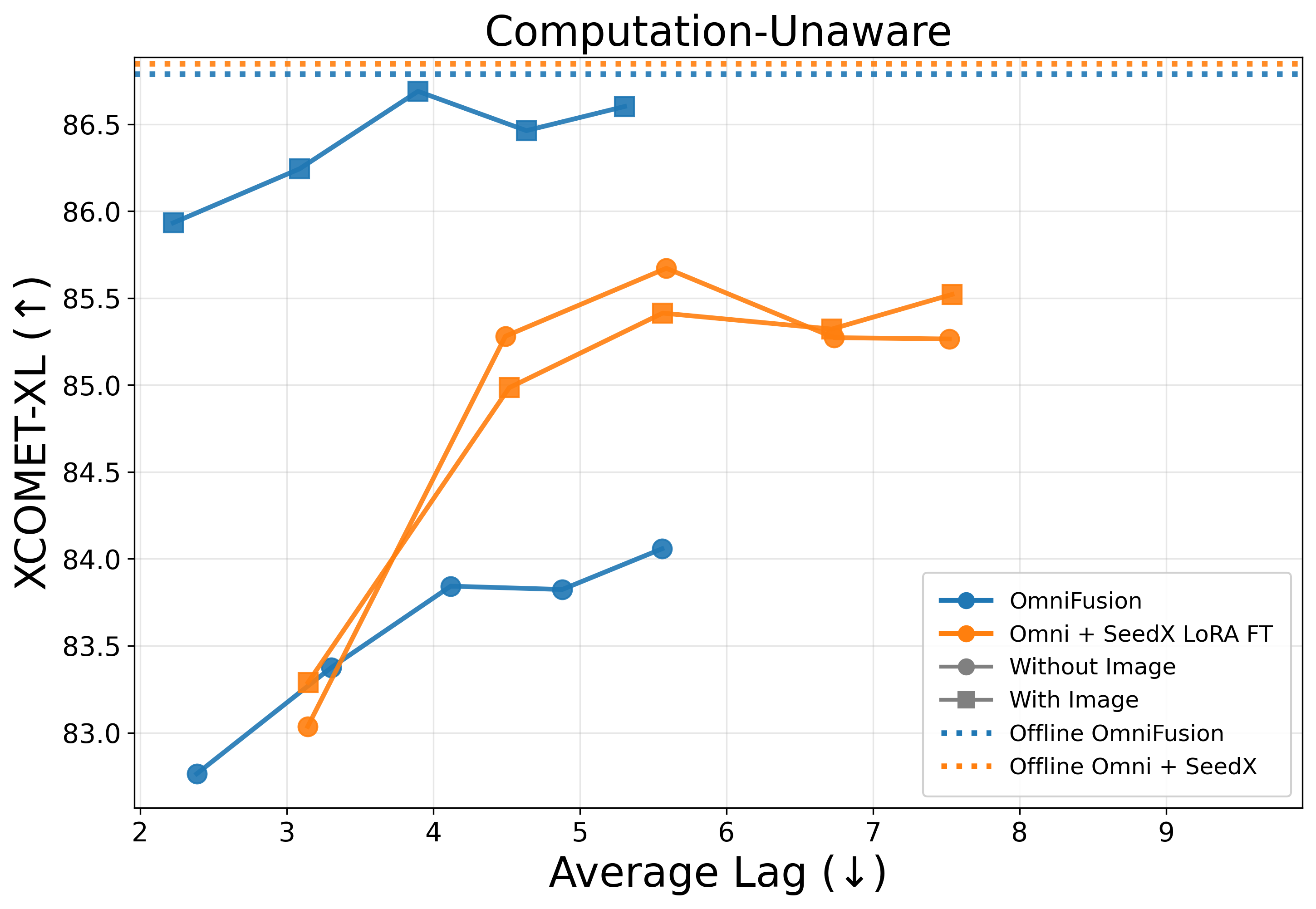}
        \caption{en $\rightarrow$ it}
        \label{fig:simul_1000_image_enit_false}
    \end{subfigure}
    }
    \vspace{1em}

    \resizebox{2\columnwidth}{!}{
    \begin{subfigure}[b]{0.45\textwidth}
        \centering
        \includegraphics[width=\textwidth]{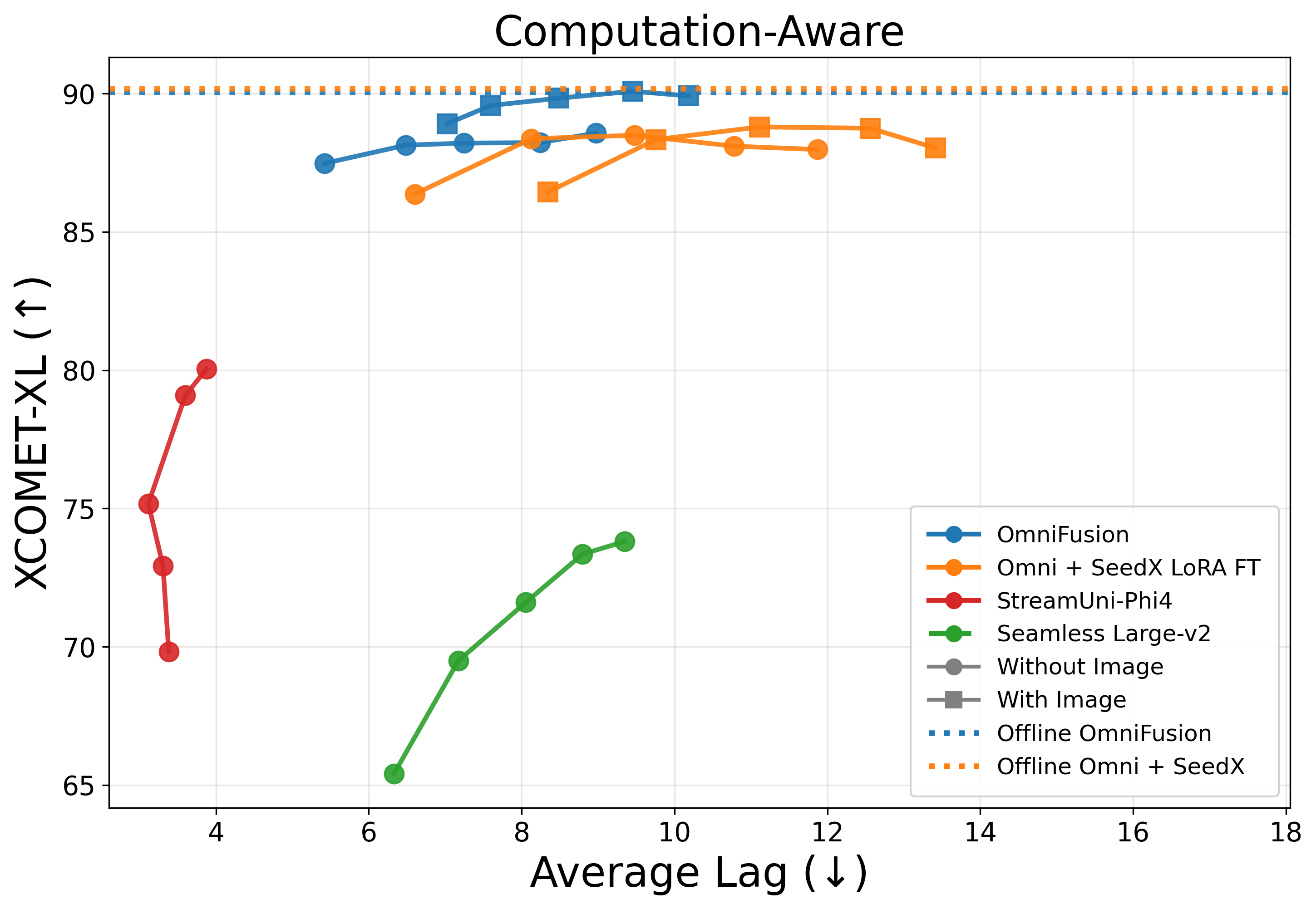}
        \caption{en $\rightarrow$ de}
        \label{fig:simul_1000_image_ende_true}
    \end{subfigure}
    \hfill
    \begin{subfigure}[b]{0.45\textwidth}
        \centering
        \includegraphics[width=\textwidth]{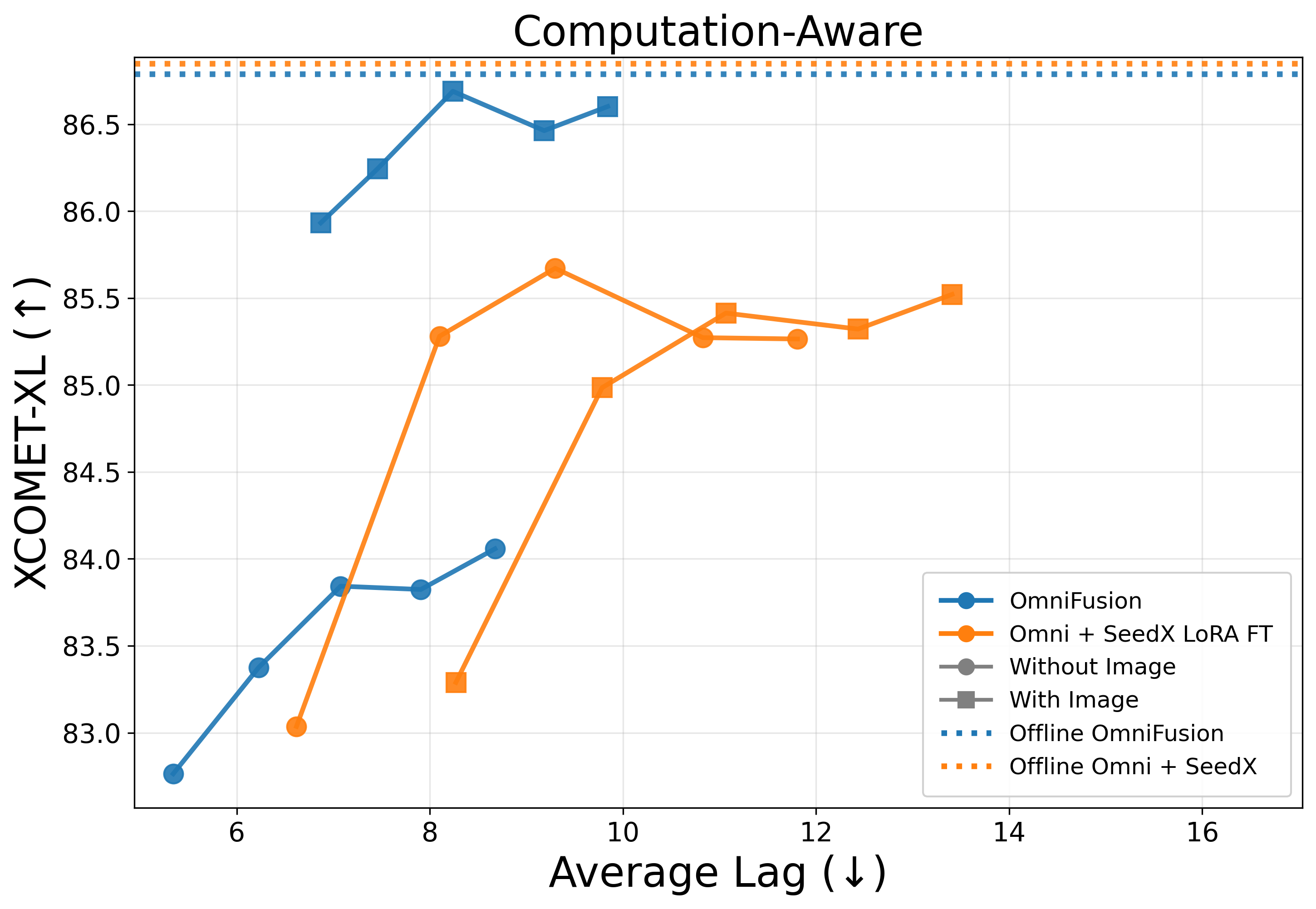}
        \caption{en $\rightarrow$ it}
        \label{fig:simul_1000_image_enit_true}
    \end{subfigure}
    }

    \caption{SimulST performance on en$\rightarrow$de and en$\rightarrow$it with chunk sizes of $[1,1.5,2,2.5,3]$ seconds. We report XCOMET-XL quality scores against Average Lagging (AL; lower is better). Top figures show computation-unaware AL, while bottom figures show computation-aware AL.}
    \label{fig:simul_st}
\end{figure*}

Our evaluation is organized around four questions. First, we ask whether OmniFusion exploiting MMFM representation improves the latency--quality trade-off for SimulST. Second, we analyze whether the proposed fusion architecture and MMFM layer selection are necessary for effective transfer to the translation LLM. Third, we examine whether the multitask training strategy is necessary, especially the OCR and self-cascading objectives. Fourth, we evaluate whether the resulting model remains competitive in offline ST and TIT, where latency constraints are removed. Finally, we perform inference-time analysis of the learned gate behavior to understand layer contribution.

\subsection{Simultaneous Speech Translation}
\label{sec:simulst}

A SimulST system must translate incrementally as speech arrives, deciding when
to read more input and when to commit target tokens. Since OmniFusion is trained
as an E2E translation model and does not natively implement streaming decoding or a learned policy,
we apply Local Agreement \citep{liu2020low} at inference time. Note that local agreement is commonly used strong strategy for SimulST in IWSLT 2025 \citep{agostinelli2025findings}. Audio is
processed in fixed-size chunks; at each step, the model generates a new
hypothesis, commits the longest common prefix between consecutive hypotheses,
and force-decodes committed tokens in subsequent steps. For the cascaded baseline, we apply the same policy in two stages: Local
Agreement is first applied to ASR hypotheses, and MT is then conditioned on the
agreed ASR prefix. This creates a strong but latency-sensitive cascade, since
translation can only proceed after ASR prefixes stabilize following IWSLT submissions \citep{agostinelli2025findings}.

We compare OmniFusion against this fine-tuned cascade built from the same pretrained
components, and include SeamlessM4T Large-v2 \citep{barrault2023seamlessm4t}  and StreamUni-Phi4 \citep{guo2025streamuni} baselines to contextualize
the latency--quality trade-off. We note that en$\rightarrow$de is the only
evaluated language pair supported by both external SimulST baselines; we do not report on en-it as it is not covered by StreamUni-Phi4 and Seamless performance was below $70$ XCOMET skewing the visualization.

We evaluate chunk sizes of $[1,1.5,2,2.5,3]$ seconds on MCIF for
en$\rightarrow$de and en$\rightarrow$it, with and without image input. We
report XCOMET-XL quality and both computation-unaware and computation-aware
Average Lagging (AL) in \Cref{fig:simul_st}; additional details are in
\Cref{sec:latency}.

The offline systems provide an expected quality upper bound, but have high latency. Under Local Agreement decoding, OmniFusion achieves better latency and quality than the fine-tuned cascade even when applying local agreement on the ASR. In the
audio-only setting ($\bullet$), it reaches comparable quality with lower latency, indicating
that the E2E model avoids part of the delay from ASR stabilization. With image
input ($\blacksquare$), OmniFusion obtains further quality gains, suggesting that visual context
helps stabilize translation decisions.

Compared with SeamlessM4T Large-v2 and StreamUni-Phi4, OmniFusion achieves
substantially higher XCOMET-XL scores in our evaluated settings.
StreamUni-Phi4 obtains lower computation-aware latency because it is designed
for streaming generation and can reuse cached computation. OmniFusion is not
trained with a streaming-specific objective; adding streaming-specific training
and caching support is a promising direction for future work. Overall, OmniFusion is able to: preserve competitive quality and reduces latency relative to the
cascade, and exploit visual context when available.

\subsection{Fusion Architecture and Layer Selection}
\label{sec:fusion_ablation}

We next analyze whether OmniFusion's performance depends on the proposed fusion design. The final model uses a gated module over three MMFM layers: the first, middle, and last layers. This design is motivated by the hypothesis that different MMFM layers encode different levels of acoustic, visual, and semantic information, and that using all layers may introduce noise rather than improve transfer.

To test this, we compare gated fusion with attention-based fusion (detailed in \Cref{sec:attention_fusion}), using either all MMFM layers or only the selected three layers. We evaluate these variants on FLEURS, including non-English source languages, to test whether the fused representations transfer beyond the English-centric training setting. Further, we also compare with Tower-Spire \citep{Ambilduke2025-kh} and SeamlessM4T v2 to test the multilingual ability.

\begin{table*}[]
\resizebox{2\columnwidth}{!}{
\begin{tabular}{@{}ccccccccccccc@{}}
\toprule
\multirow{2}{*}{Model} & \multirow{2}{*}{Mode} & \multirow{2}{*}{Fusion Layers} & \multicolumn{2}{c}{en $\rightarrow$ de} & \multicolumn{2}{c}{zh $\rightarrow$ de} & \multicolumn{2}{c}{ja $\rightarrow$ ru} & \multicolumn{2}{c}{fr $\rightarrow$ ko} & \multicolumn{2}{c}{ur $\rightarrow$ zh} \\ \cmidrule(l){4-13} 
                       &                       &                                & Chrf               & Comet-22           & Chrf               & Comet-22           & Chrf               & Comet-22           & Chrf               & Comet-22           & Chrf               & Comet-22           \\ \midrule
Tower-Spire            & Direct                & \xmark                         & 56.9               & 82.54              & N/A                & N/A                & N/A                & N/A                & N/A                & N/A                & N/A                & N/A                \\
Tower-Spire            & Self-Cascade          & \xmark                         & 61.9               & 85.02              & N/A                & N/A                & N/A                & N/A                & N/A                & N/A                & N/A                & N/A                \\
SeamlessM4T v2         & Direct                & \xmark                         & 62.4               & 85.60              & 35.8               & 70.22              & 36.9               & 71.77              & 20.8               & 74.12              & 13.5               & 61.11              \\
SeamlessM4T v2         & Self-Cascade          & \xmark                         & 63.3               & 85.81              & 36.9               & 71.34              & 38.2               & 69.85              & 21.0               & 75.22              & 13.5               & 63.75              \\ \midrule
\multicolumn{13}{c}{Our Models - OmniFusion}                                                                                                                                                                                                                                                      \\ \midrule
Attention              & Direct                & All                            & 60.9               & 84.02              & 37.4               & 72.96              & 26.6               & 65.62              & 21.2               & 75.36              & 4.3                & 43.66              \\
Attention              & Direct                & First, Mid, Last                              & 61.8               & 86.95              & 39.4               & 74.86              & 27.6               & 68.81              & 21.9               & 78.94              & 5.5                & 46.78              \\
Gated                  & Direct                & All                            & 62.3               & 85.37              & 39.6               & 75.91              & 28.8               & 69.33              & 21.8               & 77.50              & 5.0                & 45.78              \\
Gated                  & Direct                & First, Mid, Last                               & \textbf{64.0}      & \textbf{87.12}     & \textbf{44.5}      & \textbf{81.82}     & \textbf{37.2}      & \textbf{81.95}     & \textbf{25.5}      & \textbf{82.74}     & \textbf{12.5}      & \textbf{65.60}     \\ \bottomrule
\end{tabular}
}
\caption{Fusion architectures with zero-shot source languages. Tower-Spire only supports English audio.}
\label{tab:fusion_multilinguality}
\end{table*}

As shown in \Cref{tab:fusion_multilinguality}, the three-layer gated model performs best across all evaluated language pairs. Using all MMFM layers consistently degrades performance, suggesting that later MMFM representations can introduce noise when transferred to the translation LLM. Attention-based fusion is also less robust than gated fusion, particularly in zero-shot source-language settings. This indicates that a lightweight layer-wise gate is better suited for aligning MMFM representations with the translation LLM than attending over all hidden states.

Moreover, OmniFusion is better than SeamlessM4T and Tower-Spire in all directions by exploiting the state-of-the-art pretrained models. These results show that the selected-layer fusion interface is important: simply exposing the translation LLM to more MMFM layers does not improve performance. Instead, the best results in ST come from selecting a small number of transferable layers and combining them through a lightweight gating mechanism.

\subsection{Impact of Training Tasks}
\label{sec:trainingablation}

\begin{table}[!h]
\resizebox{\columnwidth}{!}{
\begin{tabular}{@{}lccc@{}}
\toprule
Training Configuration                & en$\rightarrow$de & en$\rightarrow$it & en$\rightarrow$zh \\ \midrule
W/o OCR                               & 87.12             & 82.42             & 79.75             \\
W/o (OCR + Self-Cascade)              & 89.19             & 84.35             & 82.14             \\
All Tasks                             & 89.90             & 86.79             & 82.03             \\
All Tasks (Self-Cascade Inference)    & \textbf{90.05}    & \textbf{87.20}    & \textbf{82.47}    \\ \bottomrule
\end{tabular}
}
\caption{Training task ablations on the MCIF test set.}
\label{tab:training_modes}
\end{table}

The previous ablation shows the benefit of lightweight gate, but the final
system also relies on multitask training. We therefore ablate the training tasks
from \Cref{sec:training}: one model removes OCR, and another removes both OCR
and self-cascading. We evaluate all variants on MCIF.

As shown in \Cref{tab:training_modes}, removing OCR gives the weakest
performance across all three language pairs. Interestingly, removing both OCR
and self-cascading performs better than removing OCR alone. We hypothesize that
self-cascading with images can introduce confusion when visual representations
are not well aligned, and that the OCR objective helps ground these
representations, enabling more effective self-cascade training.

Training on all tasks yields the strongest direct inference results for
en$\rightarrow$de and en$\rightarrow$it, while remaining competitive for
en$\rightarrow$zh. Under self-cascade inference, the full model performs best
across all language pairs, showing that OmniFusion's gains come not only from
fusion, but also from multitask and self-cascading objectives that better align
MMFM representations with translation generation.

\subsection{Offline Speech Translation}
\label{sec:offlinest}

Having established that both the fusion architecture and training tasks are necessary, we next evaluate the final OmniFusion system in offline ST, where the full speech input is available and latency constraints are removed. This setting allows us to compare translation quality directly against cascaded systems and to assess whether the E2E model remains competitive when the cascade is not penalized for latency.

\begin{table*}[t!]
\resizebox{2\columnwidth}{!}{
\centering
\small
\setlength{\tabcolsep}{6pt}
\begin{tabular}{@{}c|ccccccc@{}}
\toprule
\textbf{Category} & \textbf{Config} & \textbf{Image} & \textbf{XCOMET-XL} & \textbf{Minor} & \textbf{Major} & \textbf{Critical} & \textbf{Total} \\ \midrule
\multirow{1}{*}{\textbf{\begin{tabular}[c]{@{}c@{}}Omni FT (Direct ST)\\\end{tabular}}} 
& LoRA FT          & \xmark     & 82.30    & \textbf{1079.0}  & 852.0   & 90.3 & 2021.3 \\ \midrule
\multirow{4}{*}{\textbf{\begin{tabular}[c]{@{}c@{}}Cascade Systems (ASR $\rightarrow$ MT)\\ Omni + SeedX\end{tabular}}} 
& No FT            & \xmark         & 85.88             & 1212.3 & 773.0 & 74.3 & 2059.7 \\
& No FT            & \checkmark     & 85.94             & 1191.3 & 759.3 & 74.7 & 2025.3 \\
& LoRA FT       & \xmark         & 86.42             & 1239.0 & 751.3 & 75.0 & 2065.3 \\
& LoRA FT       & \checkmark     & \textbf{86.59}    & 1202.3 & 766.3 & 55.7 & 2024.3 \\ \midrule
\multirow{5}{*}{\textbf{\begin{tabular}[c]{@{}c@{}}End-to-End (Direct ST)\\ OmniFusion\end{tabular}}}
& Mid Fusion            & \xmark         & 83.57             & 1114.0 & 1188.3 & 72.3 & 2374.7 \\
& Mid Fusion            & \checkmark     & 85.98             & 1256.3 & 749.0 & 61.7 & 2067.0 \\
& Gated Fusion         & \xmark         & 83.98             & 1119.3 & 1188.0 & 62.7 & 2370.0 \\
& Gated Fusion           & \checkmark     & 86.24             & 1255.7 & 736.0 & 57.0 & 2048.7 \\
& Gated Fusion  + Self-Cascade & \checkmark & 86.57           & 1231.0 & \textbf{719.0} & \textbf{55.3} & \textbf{2005.3} \\ \bottomrule
\end{tabular}
}
\caption{Offline ST results averaged across en$\rightarrow$zh, en$\rightarrow$de, and en$\rightarrow$it on the MCIF test. \checkmark\ indicates models using image input. Bold indicates best XCOMET-XL and lowest error counts. Language-wise results are additionally reported in Table \ref{tab:en_de_xcomet}, \ref{tab:en_it_xcomet}, and \ref{tab:en_zh_xcomet}.}
\label{tab:averaged_xcomet}
\end{table*}

\begin{table*}[!ht]
\centering
\begin{tabular}{@{}c|c|cccccc@{}}
\toprule
Model & Config & en $\rightarrow$ ar & en $\rightarrow$ cs & en $\rightarrow$ de & en $\rightarrow$ fr & en $\rightarrow$ ru & en $\rightarrow$ zh \\ 
\midrule
Omni 2.5 7B & - & 78.56 & 81.03 & 81.94 & 79.99 & 47.27 & 81.74 \\
Seed-X-PPO-7B & - & \textbf{80.39} & 83.56 & 82.38 & 80.31 & 82.13 & 83.25 \\
ZeroMMT 3.3B & - & \textbf{80.64} & \textbf{86.69} & 83.54 & 83.40 & 83.87 & 77.85 \\
TowerVision 9B & - & 70.85 & 86.62 & \textbf{84.97} & \textbf{84.29} & \textbf{85.81} & \textbf{85.24} \\
OmniFusion 14B & Mid Fusion & \textbf{80.98} & \textbf{87.49} & \textbf{85.14} & \textbf{84.97} & \textbf{85.69} & \textbf{85.15} \\
OmniFusion 14B & Gated Fusion & \textbf{80.81} & \textbf{87.58} & \textbf{84.88} & \textbf{84.28} & 84.62 & 84.34 \\
\bottomrule
\end{tabular}
\caption{COMET scores (↑) on the CoMMuTE dataset. Statistically significant top-performing results per language pair are highlighted in \textbf{bold} (computed using \texttt{comet-compare} with default settings). Qualitative example can be found in \Cref{tab:commute_example}.}
\label{tab:commute_comet}
\end{table*}

\Cref{tab:averaged_xcomet} reports XCOMET-XL scores and fine-grained error counts on MCIF, averaged over en$\rightarrow$zh, en$\rightarrow$de, and en$\rightarrow$it. Directly fine-tuning the MMFM for speech translation performs substantially worse than OmniFusion, confirming that adapting the MMFM alone is insufficient. In the cascaded pipeline, LoRA fine-tuning improves quality, while image input provides only marginal gains. In contrast, OmniFusion benefits more clearly from image context, suggesting that our model can use visual information more directly than the cascade.

The best cascaded system achieves the highest aggregate XCOMET-XL score, showing that cascades remain strong when full input and latency are not constraints. However, OmniFusion with gated fusion, image input, and self-cascade inference nearly matches it while producing fewer severe errors. This suggests the end-to-end fused model improves not only latency but also robustness by conditioning directly on multimodal context rather than only an ASR transcript.

\subsection{Text-Image Translation}
\label{sec:tit}

We further evaluate whether OmniFusion can exploit visual context beyond speech translation. CoMMuTE targets visually grounded ambiguity, making it a useful benchmark for assessing lexical and semantic disambiguation with image context. As shown in \Cref{tab:commute_comet}, the unimodal translation model SeedX performs competitively, but is generally weaker than multimodal systems. OmniFusion improves over the individual base components and matches or surpasses strong multimodal baselines such as ZeroMMT and TowerVision on most language pairs. We also conduct a small manual disambiguation evaluation in \Cref{sec:commute}. TowerVision performs better in this targeted setting, but OmniFusion remains competitive despite not being extensively trained for image-only translation and while additionally supporting audio input. This suggests that fusing a pretrained MMFM with a translation LLM can yield strong visually grounded translation without training from scratch.

Mid Fusion slightly outperforms Gated Fusion on CoMMuTE, indicating that fusion should match the modality’s role. Middle-layer fusion is sufficient for text-image translation, where images mainly disambiguate text, while gated fusion is better suited to speech translation, where audio is the primary input. Thus, MMFM layer selection matters, and no fusion strategy is universally optimal.

\subsection{Inference-Time Gate Behavior}
\begin{figure}[!hb]
    \centering
    \includegraphics[width=\columnwidth]{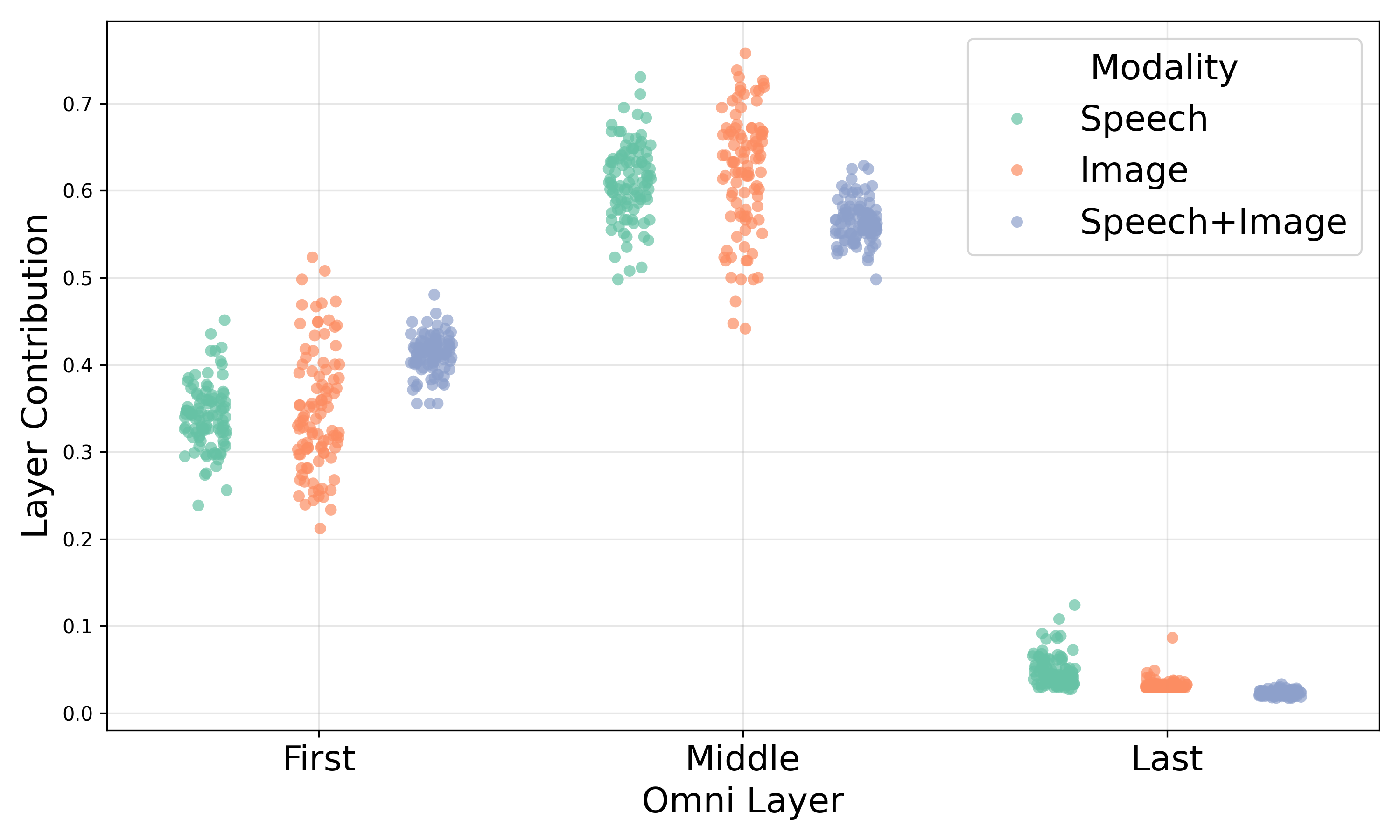}
    \caption{Gate predictions across modalities.}
    \label{fig:layer_contrib}
\end{figure}

Finally, we analyze how the trained gate uses MMFM layers across different
input settings. We compute the average gate weights assigned to the first, middle, and last
MMFM layers for speech-only, image-only, and speech+image inputs. As shown in
\Cref{fig:layer_contrib}, the gate consistently assigns low weight to the last
layer, suggesting that final MMFM states are less useful as transfer points for
translation. The relative use of the first and middle layers varies by task:
speech inputs place more weight on early representations, while image-only and
speech+image inputs rely more strongly on middle-layer representations. This
indicates that the gate adapts to the available modality rather than selecting a
fixed layer for all tasks.

We see that useful transfer
depends not only on connecting an MMFM to a translation LLM, but also on
selecting modality-appropriate internal representations.


\section{Related Work}

\textbf{Multimodal MT}: Caption translation has been widely studied
\citep{specia-etal-2016-shared, elliott-etal-2017-findings,
barrault-etal-2018-findings}, including work that trains multimodal NMT models
from scratch \citep{vijayan-etal-2024-adding} or augments pretrained MT models
such as NLLB \citep{costa2022no,futeral-etal-2025-towards}. Recent work extends
translation LLMs to images, video, or speech
\citep{Viveiros2025-dc,Ambilduke2025-kh}, while fewer studies jointly use
speech and visual context \citep{Sinhamahapatra2025-cf,anwar2023muavic,
choi2024av2av}. In contrast, OmniFusion jointly leverages images and audio
through a lightweight fusion interface.

\textbf{Simultaneous ST}: Real-time ST requires translating before the full
utterance is available. Prior work often adapts offline models using fixed or
learned policies such as wait-$k$ \citep{ma-etal-2019-stacl,
elbayad2020efficient}, Local Agreement \citep{liu2020low}, attention-based
heuristics \citep{papi2023alignatt}, or LLM-based policies
\citep{koshkin-etal-2024-transllama, koshkin-etal-2024-llms,
guo2025streamuni, fu2025efficient}. We use Local Agreement for evaluation, but
OmniFusion is compatible with other SimulST policies.

\textbf{Modality Fusion:} Multimodal LLMs typically process modalities with
separate encoders and merge them in the LLM
\citep{Abdin2024-ez, Xu2025-fv, gemmateam2025gemma3technicalreport}, while
other models enable earlier audio-visual interaction \citep{ye2025omnivinci}.
Related work also studies modality projectors \citep{verdini2024connect,
gaido2024speech} and discrete tokens \citep{zhan-etal-2024-anygpt,
li2025discrete}. OmniFusion instead studies which internal MMFM layers provide
transferable representations.

\section{Conclusion}

We introduced OmniFusion, a modular framework that connects a frozen MMFM with a translation LLM for speech, speech-image, and text-image translation. Experiments show that early and middle MMFM layers transfer better than late layers, and that selected-layer fusion improves over all-layer fusion in ST. OmniFusion improves the SimulST latency--quality trade-off under Local Agreement decoding while remaining competitive in offline ST and CoMMuTE. Finally, the best fusion strategy is task-dependent: gated fusion is more robust for ST, while middle-layer fusion can suffice when images mainly provide disambiguation context.

\section*{Limitations}

Our training data is primarily English-centric, which may limit performance for other source languages. However, our zero-shot experiments on non-English source languages suggest encouraging transfer, but broader source-language training and evaluation are needed. Another limitation of our work is that OmniFusion is trained with fixed task-specific prompts across the three supported translation settings, and therefore does not exhibit general instruction-following capabilities. It remains unclear whether our fusion strategy can be extended to support more flexible, instruction-driven multimodal behavior. Our manual CoMMuTE analysis is small-scale and shows that  OmniFusion achieves reasonable performance on targeted visual disambiguation. This suggests that broader human evaluation is needed to fully assess visually grounded translation behavior.

\section*{Acknowledgments}
The research leading to these results was supported
by European Union’s Horizon Europe programme
grant agreement No. 101213369 (DVPS).

\bibliography{ref}
\appendix

\section{Appendix}
\label{sec:appendix}

\subsection{CoMMuTE Manual Evaluation}
\label{sec:commute}

To further validate whether the model is genuinely disambiguating examples in
the CoMMuTE benchmark, we manually annotate the \textbf{first 50 source sentences from
the English-to-German direction}, each paired with two images, resulting in a
subset of 150 examples. We compare OmniFusion against TowerVision on this
subset. The annotation is performed by a single annotator who inspects the
source sentence together with the corresponding image. To reduce potential bias,
we provide the annotator with the hypothesis files together with the
disambiguation labels. For each example, we evaluate whether the system
correctly disambiguates the sentence under both image conditions, while
ignoring overall translation quality\footnote{The source and reference sentences along with model hypothesis are submitted in data upload.}.

We find that, out of 50 source sentences, OmniFusion correctly disambiguates
18 examples, whereas TowerVision correctly disambiguates 23. Although the
absolute scores are modest, the benchmark consists of highly challenging and
synthetic examples. Nonetheless, it is still better than a text-only baseline as it will not disambiguate all and have a score of $0$.

Moreover, while OmniFusion underperforms TowerVision on
this targeted disambiguation analysis, it is important to note that OmniFusion
also supports the audio modality and was not trained on the same scale of data
as TowerVision.

This suggests that, despite lower performance on this image-focused
disambiguation subset, OmniFusion retains broader multimodal applicability
while requiring substantially less modality-specific training data.

\subsection{Inference-Time Gate Control}
\label{sec:gate_analysis}

\begin{table}[!ht]
\resizebox{\columnwidth}{!}{
\begin{tabular}{@{}ccc@{}}
\toprule
Inference-time Gate Force & Speech       & Speech + Image \\ \midrule
First      & 55.22            & 63.02             \\
Middle     & Halluc. + ASR & Halluc. + ASR  \\
Last       & EOS       & EOS        \\
\xmark   & 89.90         & 90.05          \\ \bottomrule
\end{tabular}
}
\caption{Inference-time control of the gating module by forcing the first, middle, or last MMFM layer. We report XCOMET-XL scores on the MCIF en $\rightarrow$ de test set.}
\label{tab:inference}
\end{table}

To further test the gate, we force the model to use only one MMFM layer at inference time and report results in \Cref{tab:inference}. Forcing the first layer degrades quality but does not collapse the system, suggesting that early MMFM representations remain partially aligned with the translation LLM. In contrast, forcing the middle layer leads to hallucinated transcripts and ASR-like outputs, while forcing the last layer causes the model to emit only the \texttt{eos} token. These failure modes show that the gate does more than select a single best layer: it balances complementary information from multiple MMFM depths.

Together, the ablations and analysis support the main claim of OmniFusion. The model requires both an effective architectural interface and an appropriate training strategy: selected MMFM layers and gated fusion provide transferable multimodal representations, while OCR and self-cascading objectives make those representations useful for translation.

\subsection{Attention Fusion}
\label{sec:attention_fusion}

\paragraph{Attention-based layer fusion.}
As an alternative to the gated fusion module, we also consider a
token-wise attention-based fusion mechanism over the selected MMFM layers.
We evaluate this attention-based fusion using the first, middle, and last
layers, and further analyze its behavior when applied to other selected layer
sets. Below, we formalize the method using three layers as an example, although
the formulation can be extended to an arbitrary number of selected layers.

For each multimodal token $z_t$, we first stack the three selected layer
representations into a length-three layer sequence:
\begin{equation}
\label{eq:attn_stack}
\begin{aligned}
H_t
&=
\left[
h_t^{(\mathrm{first})};
h_t^{(\mathrm{mid})};
h_t^{(\mathrm{last})}
\right], \\
H_t
&\in
\mathbb{R}^{3 \times D_{\mathrm{MMFM}}},
\qquad
t = 1,\dots,L_m .
\end{aligned}
\end{equation}
Let $\mathcal{J}=\{\mathrm{first},\mathrm{mid},\mathrm{last}\}$ denote the set
of selected layers. Then $h_t^{(j)}$ denotes the representation of token $z_t$
from layer $j$, where $j \in \mathcal{J}$.

We compute a token-wise query vector from the mean of the selected layer
representations:
\begin{equation}
\label{eq:attn_query}
\begin{aligned}
\bar{h}_t
&=
\frac{1}{|\mathcal{J}|}
\sum_{j \in \mathcal{J}} h_t^{(j)}, \\
q_t
&=
W_Q \bar{h}_t + b_Q, \\
q_t
&\in
\mathbb{R}^{D_{\mathrm{MMFM}}},
\end{aligned}
\end{equation}
where $W_Q \in \mathbb{R}^{D_{\mathrm{MMFM}} \times D_{\mathrm{MMFM}}}$
and $b_Q \in \mathbb{R}^{D_{\mathrm{MMFM}}}$ are learnable parameters.

The attention score for each selected layer is computed by a dot product
between the query and the corresponding layer representation:
\begin{equation}
\label{eq:attn_scores}
\begin{aligned}
s_t^{(j)}
&=
q_t^\top h_t^{(j)},
\qquad
j \in \mathcal{J}.
\end{aligned}
\end{equation}
The scores are normalized across the selected layers:
\begin{equation}
\label{eq:attn_weights}
\begin{aligned}
\beta_t^{(j)}
&=
\frac{\exp(s_t^{(j)})}
{\sum_{\ell \in \mathcal{J}} \exp(s_t^{(\ell)})},
\qquad
j \in \mathcal{J}, \\
\beta_t
&=
\left[
\beta_t^{(j)}
\right]_{j \in \mathcal{J}}
\in
\mathbb{R}^{|\mathcal{J}|}.
\end{aligned}
\end{equation}
The final attention-fused MMFM representation is obtained as a weighted sum
over the selected layer representations:
\begin{equation}
\label{eq:attn_fused_mmfm}
\begin{aligned}
r_t
&=
\sum_{j \in \mathcal{J}}
\beta_t^{(j)} h_t^{(j)}, \\
r_t
&\in
\mathbb{R}^{D_{\mathrm{MMFM}}},
\qquad
t = 1,\dots,L_m .
\end{aligned}
\end{equation}
The fused representation $r_t$ is then projected into the translation-model
embedding space using the same projection MLP:
\begin{equation}
\label{eq:attn_mmfm_projection}
\begin{aligned}
v_t
&=
\mathrm{MLP}(r_t), \\
v_t
&\in
\mathbb{R}^{D_{\mathrm{NMT}}},
\qquad
t = 1,\dots,L_m .
\end{aligned}
\end{equation}

\begin{table*}[!h]
\resizebox{2\columnwidth}{!}{
\centering
\begin{tabular}{ccccccccc}
\hline
\multirow{2}{*}{\textit{Model}} & \multirow{2}{*}{\textit{Mode}} & \multirow{2}{*}{\textit{Hypothesis}} & \multicolumn{6}{c}{\textit{Comet-22 ($\uparrow$)}}                                                                               \\ \cline{4-9} 
                                &                                &                                      & en $\rightarrow$ de & en$\rightarrow$ zh & \underline{zh} $\rightarrow$ de & \underline{es} $\rightarrow$ it & \underline{ja} $\rightarrow$ \underline{ru} & \underline{fr} $\rightarrow$ \underline{ko} \\ \hline
\multicolumn{9}{c}{\textit{\textbf{Baselines}}}                                                                                                                                                                                            \\ \hline
Qwen2 Audio                           & E2E                            & \_                                   & 81.24                   & 78.20                  & 73.59                   & 83.38                   & 70.70                   & 71.26                   \\
Qwen2 Audio  FT                       & E2E                            & \_                                   & 84.82                   & 85.57                  & 80.22                   & 84.17                   & 67.69                   & 77.09                   \\
Qwen2 Audio + Tower                   & Cascade                        & \_                                   & 85.85                   & \textbf{85.97}                  & 82.94                   & 85.36                   & 69.26                   & \textbf{84.78}                   \\ 
Tower APE                   & Cascade APE                        & Qwen2 Audio + Tower                                   & 85.96                   & \textbf{85.77}                  & 82.59                   & 85.21                   & 47.43                   & 70.67                   \\ \hline
\multicolumn{9}{c}{\textit{\textbf{Off-the-shelf ST models}}}                                                                                                                                                                                       \\ \hline
Tower-Spire                     & E2E                            & \_                                   & 82.77                   & 81.63                  & \_                   & \_                   & \_                   & \_                   \\
Tower-Spire                     & Self-Cascade                   & \_                                   & 85.17                   & 85.73                  & \_                  & \_                   & \_                   & \_                   \\
SeamlessM4T v2                 & E2E                            & \_                                   & 85.63                   & 80.05                      & 70.22                   & 77.78                   & 70.58                   & 78.91                   \\ \hline
\multicolumn{9}{c}{\textit{\textbf{E2E models}}}                                                                                                                                                                                       \\ \hline
AudioTower-Zero                         & E2E                            & \_                                   & 85.13                   & 83.97                  & 76.62                   & 83.98                   & 68.71                   & 81.64                   \\
AudioTower-Mid                          & E2E                            & \_                                   & 86.02                   & 85.03                  & 80.62                   & 84.86                   & 74.47                  & 83.26                   \\
AudioTower-Last                         & E2E                            & \_                                   & 82.97                   & 81.03                  & 54.12                   & 49.57                   & 50.19                   & 52.31                   \\ \hline
\multicolumn{9}{c}{\textit{\textbf{E2E + Speech-Aware APE model}}}                                                                                                                                                                     \\ \hline
AudioTower-Mid                          & Cascade APE                    & Qwen2 Audio + Tower                        & \textbf{86.34}                    & \textbf{85.99}                   & \textbf{83.42}                    & \textbf{85.96}                    & \textbf{78.26}                    & \textbf{84.59}                    \\
AudioTower-Mid                          & E2E                    & Empty                                & 85.77                    & 84.79                   & 80.39                    & 85.27                    & 74.67                    & 82.71                    \\ \hline
\end{tabular}
}
\caption{Offline ST results on FLEURS test set. Speech-Aware APE models use the hypothesis from a system and perform post-editing. Statistically significant top-performing results per language pair are highlighted in \textbf{bold}. \underline{Underlined} languages are not seen during the fusing process as source/target.}
\label{tab:layer_analysis}
\end{table*}

\begin{table}[]
\resizebox{\columnwidth}{!}{
\begin{tabular}{@{}ccccc@{}}
\toprule
\textbf{Dataset} & \textbf{\# Examples} & \textbf{Language} & \textbf{Modality} & \textbf{Task} \\ \midrule

M3AV & 122k & \textit{en} & Speech, Image & ASR, OCR \\ \midrule

M3AV - Aug & 122k &
\begin{tabular}[c]{@{}c@{}}
\textit{zh}, \textit{fr}, \textit{de}, \textit{it}, \\ \textit{ja}, \textit{ko},
\textit{pt}, \textit{ru}, \\ \textit{es}, \textit{vi}, \textit{ar}, \textit{cs},\\
\textit{hr}, \textit{da}, \textit{nl}, \textit{fi}, \\ \textit{hu}, \textit{id},
\textit{ms}, \textit{nb}, \\ \textit{no}, \textit{pl}, \textit{ro}, \textit{tr}
\end{tabular}
& Speech, Image & ST, OCR \\ \midrule

Europarl & \multirow{2}{*}{260k*} &
\begin{tabular}[c]{@{}c@{}}
\textit{de}, \textit{es}, \textit{fr}, \textit{it}, \textit{nl},\\
\textit{pl}, \textit{pt}, \textit{ro}
\end{tabular}
& Speech & ST \\

\cmidrule(r){1-1} \cmidrule(l){3-5}

Covost2 & & 
\textit{de}, \textit{ja}, \textit{tr}, \textit{zh}
& Speech & ST \\ \midrule

Multi30k & 87k &
\textit{cs}, \textit{de}, \textit{fr}
& Image & TIT \\

\bottomrule
\end{tabular}
}
\caption{Overview of data sources collected for E2E training. \textbf{*} We sample 260k examples uniformly per language and dataset (~260k / 24) , with German appearing more frequently due to its presence in both datasets. All the speech and text data are only available with English as the source language. Language code mapping is presented in \Cref{tab:lang-codes}.}
\label{tab:data_overview}
\end{table}

\begin{table}[t]
\centering
\resizebox{\columnwidth}{!}{
\begin{tabular}{l c c c}
\toprule
\textbf{Model} & \textbf{Time(s)} & \textbf{Image} & \textbf{XCOMET-XL} \\
\midrule

Omni + SeedX FT  & 2.96 & w/o  & 89.75 \\
OmniFusion       & \textbf{1.98} & w/o  & 88.09 \\
OmniFusion SC & 3.78 & w/o  & 88.44 \\

\midrule

Omni + SeedX FT  & 3.85 & w/  & \textbf{90.18} \\
OmniFusion       & 3.15 & w/  &  89.90 \\
OmniFusion SC & 4.82 & w/  &  90.05 \\
\bottomrule
\end{tabular}
}
\caption{Quality vs latency tradeoffs in the offline setting for English $\rightarrow$ German on the MCIF test set. SC indicates Self-Cascade mode.}
\label{tab:quality_latency}
\end{table}

\begin{table*}[!ht]
\resizebox{2\columnwidth}{!}{
\centering
\begin{tabular}{@{}ccccccccc@{}}
\toprule
\multirow{2}{*}{\textit{Model}} & \multirow{2}{*}{\textit{Mode}} & \multirow{2}{*}{\textit{Hypothesis}} & \multicolumn{6}{c}{\textit{Comet-22 ($\uparrow$)}}                                                                               \\ \cmidrule(l){4-9} 
                                &                                &                                      & en $\rightarrow$ de & en$\rightarrow$ zh & \underline{zh} $\rightarrow$ de & \underline{es} $\rightarrow$ it & \underline{ja} $\rightarrow$ \underline{ru} & \underline{fr} $\rightarrow$ \underline{kr} \\ \midrule
Qwen2 Audio + Tower                   & Cascade                        & \_                                   & 85.85               & 85.97              & 82.94               & 85.36               & 69.26               & 84.78               \\
Whisper + Tower                 & Cascade                        & \_                                   & 86.71               & \textbf{86.31}     & 82.23               & \textbf{86.54}      & 83.56               & \textbf{85.29}      \\
AudioTower-Mid                          & Cascade APE                    & Qwen2 Audio + Tower                        & 86.34               & 85.99              & \textbf{83.42}      & 85.96               & 78.26               & 84.59               \\
AudioTower-Mid                          & Cascade APE                    & Whisper + Tower                      & \textbf{87.09}      & \textbf{86.34}     & 82.81               & \textbf{86.60}      & \textbf{84.38}      & 84.95               \\ \midrule
\multicolumn{9}{c}{Character-Error-Rate ($\downarrow$)}                                                                                                                                                                                    \\ \midrule
Whisper                         & ASR                            & \_                                   & 3.14                & 3.15               & 25.37               & 2.11                & 7.67                & 3.28                \\ \bottomrule
\end{tabular}
}
\caption{Analysis of system combination by using Whisper to generate the hypothesis for post-editing on FLEURS test set. Statistically significant top-performing results per language pair are highlighted in \textbf{bold}. \underline{Underlined} languages are not seen during the fusing process as source/target.}
\label{tab:system_comb}
\end{table*}

\subsection{Prompt Building}

\begin{tcolorbox}[promptbox, title=System Prompt for all modalities]
You are Qwen, a virtual human developed by the Qwen Team, Alibaba Group, capable of perceiving auditory and visual inputs, as well as generating text and speech.
\end{tcolorbox}

\begin{tcolorbox}[promptbox, title=User Prompt (ST with/without Image)]
Transcribe the audio using the image for context, and perform OCR on the image for correct spelling of keywords and names.
\end{tcolorbox}

\begin{tcolorbox}[promptbox, title=User Prompt (TIT)]
Translate the following sentence into \texttt{\{tgt\_lang\}} using the image for context.  
\texttt{<IMG>} refers to the image.  
Use the image to determine formality, gender, keywords, and other details important for disambiguation:

\texttt{\{ex[src]\}} \texttt{<IMG>}
\end{tcolorbox}

\begin{algorithm*}[p]
\label{alg:multimodal}
\begin{algorithmic}[1]
\Require Example $ex$
\Statex \textit{// SeedX prompt (source text, language tag and reference)}
\State $\texttt{src}, \texttt{tgt}, \texttt{img}, \texttt{aud}, \texttt{ocr}, \texttt{lang}
        \gets \Call{Extract}{ex}$
\State $\texttt{suffix} \gets \langle$\textsc{Lang}$(\texttt{lang})\rangle$
\Statex

\If{$\texttt{aud}$ \textbf{ and not } $\texttt{img}$} \Comment{Speech only}
    \If{$\Call{Random}{} < 0.1$}
        \State $\texttt{inp} \gets \langle$\texttt{en}$\rangle$ \Comment{English-Audio only in our dataset}
        \State $\texttt{lbl} \gets \Call{ASR}{\texttt{src}} 
                {+}\langle$\texttt{eos}$\rangle{+}\texttt{suffix}{+}\texttt{tgt} 
                {+}\langle$\texttt{eos}$\rangle$ \Comment{ASR Self-cascade}
    \Else
        \State $\texttt{inp} \gets \texttt{suffix}$
        \State $\texttt{lbl} \gets \texttt{tgt} {+} \langle$\texttt{eos}$\rangle$
    \EndIf

\ElsIf{$\texttt{img}$ \textbf{ and not } $\texttt{aud}$} \Comment{Image only}
    \If{$\Call{Random}{} < 0.05$}
        \State \Call{Swap}{$\texttt{src},\texttt{tgt}$}
        \State $\texttt{suffix} \gets \langle$\texttt{English}$\rangle$
    \EndIf
    \State $\texttt{src}_m \gets \Call{Mask}{\texttt{src}, \texttt{prob\_words}=0.2, \texttt{mask\_chance}=0.2}$ \Comment{Mask 20\% source words for 20\% of the time to pay attention to Image}
    \State $\texttt{inp} \gets \langle$\textsc{Src}$\rangle\;\texttt{src}_m
            {+}\langle$\texttt{eos}$\rangle{+}\texttt{suffix}$
    \State $\texttt{lbl} \gets \texttt{tgt} {+} \langle$\texttt{eos}$\rangle$

\ElsIf{$\texttt{img}$ \textbf{ and } $\texttt{aud}$} \Comment{Audio + Image}
    \If{$\Call{Random}{} < 0.8$}
        \If{$\Call{Random}{} < 0.1$}
            \State $\texttt{inp} \gets \langle$\texttt{en}$\rangle$
            \State $\texttt{lbl} \gets \Call{ASR}{\texttt{src}}
                    {+}\langle$\texttt{eos}$\rangle{+}\texttt{suffix}
                    {+}\texttt{tgt}{+}\langle$\texttt{eos}$\rangle$
        \Else
            \State $\texttt{inp} \gets \texttt{suffix}$
            \State $\texttt{lbl} \gets \texttt{tgt} {+} \langle$\texttt{eos}$\rangle$
        \EndIf
    \Else
        \State $\texttt{inp} \gets \langle$\textsc{Ocr}$\rangle$ \Comment{OCR Self-cascade}
        \State $\texttt{lbl} \gets \texttt{ocr}
                {+}\langle$\texttt{eos}$\rangle{+}\texttt{suffix}
                {+}\texttt{tgt}{+}\langle$\texttt{eos}$\rangle$
    \EndIf

\Else
    \State \textbf{error} ``no modalities''
\EndIf
\Statex
\Statex \textit{// Omni prompt (Multimodal prompt builder)}
\If{$\texttt{aud}$ \textbf{ and } $\texttt{img}$}
    \State $\texttt{user} \gets \texttt{``Transcribe with OCR and image context:''} \;\Vert\;
           \texttt{img} \;\Vert\; \texttt{aud}$
\ElsIf{$\texttt{aud}$}
    \State $\texttt{user} \gets \texttt{``Transcribe audio:''} \;\Vert\; \texttt{aud}$
\Else
    \State $\texttt{user} \gets \texttt{``Translate with image:''} \;\Vert\; \texttt{img}$
\EndIf
\State \Call{Store}{$\texttt{user}$,$\texttt{inp},\texttt{lbl}$}
\end{algorithmic}
\caption{Online Prompt building for examples in the training data to enable alignment and self-cascading inference mode.}
\label{alg:prompt}
\end{algorithm*}

\subsection{Layer Analysis, Zero-shot ability and APE - AudioTower}

In our preliminary experiments, we focus exclusively on the audio modality and investigate fusing Qwen-Audio 7B \citep{chu2023qwen} with Tower-Instruct 7B v0.2 \citep{Alves2024-ls}. We train the fused model for ST using the same setup described earlier, but without gating, and we independently fuse the first, middle, and last layers of Qwen-Audio to assess their individual contributions. We then compare this fused model against several baselines. First, we evaluate both the out-of-the-box and fine-tuned versions of Qwen-Audio (Qwen2 Audio FT), trained end-to-end on the same ST data. We also include a standard cascaded setup using off-the-shelf ASR and MT models. Additionally, we evaluate an Automatic Post-Editing (APE) configuration in which Tower corrects translations generated by Qwen-Audio.

To contextualize performance, we further compare against other off-the-shelf ST systems such as Tower-Spire \citep{Ambilduke2025-kh} and SeamlessM4T \citep{barrault2023seamlessm4t}, which provides insight into multilingual generalization. Beyond the main fused model, we also train an APE variant of our fusion architecture that post-edits the outputs of cascaded systems. Unlike traditional APE, this Speech-Aware APE allows the model to use the source audio when correcting errors, enabling more informed revisions. For both APE variants, we generate synthetic APE tuples by running Qwen-Audio on the training data and fine-tune with LoRA.

We report results on the FLEURS test set in Table~\ref{tab:layer_analysis}, including evaluations on zero-shot language directions not present on either the source or target side during training. First, while fine-tuning improves Qwen2 Audio, it still falls short of the out-of-the-box cascaded setup across all languages, underscoring the strength of translation-centric LLMs. Tower-based APE is competitive for seen languages but struggles in harder zero-shot directions, such as \textit{fr}→\textit{kr}. Off-the-shelf end-to-end ST models lag significantly behind cascaded systems as well.

Our proposed fused model (AudioTower) with middle-layer fusion achieves the strongest overall performance and even surpasses the cascaded baseline in several languages. Notably, both the first and middle layers show strong transfer to unseen languages, whereas the last layer collapses entirely—supporting the hypothesis that upper layers encode highly language-specific features, making them less suitable for cross-lingual contextual embeddings.

Finally, the APE variant of the fused model delivers the best results overall by combining the strong hypotheses of the cascaded pipeline with the ability to exploit the raw audio when correcting errors. We also test this APE model with empty hypotheses and observe performance comparable to AudioTower-Mid, indicating its flexibility for multiple use cases.

Given the strong APE results, we additionally explore whether the model can correct outputs from different ASR systems. Using Whisper \citep{radford2023robust} for ASR, we evaluate the fused APE model on Whisper+Tower outputs and report results in \Cref{tab:system_comb}. When Whisper+Tower produce the strongest cascaded baseline, using them as hypotheses likewise yields the highest APE performance—highlighting that Speech-Aware APE can be beneficial not only for E2E setups but also for offline cascaded scenarios.

\subsection{Latency Reproduction}
\label{sec:latency}

We use the SimulST toolkit to calculate Average Lagging for both computation-aware and computation-unaware settings. For a given chunk size, we iteratively feed the system audio segments of that duration and apply the local agreement algorithm. At each step, we commit any finalized words and record both the committed output and the amount of audio consumed.

In the E2E setup, local agreement is applied directly to the translation output. In the cascade setup, however, we first wait for the ASR output to stabilize before applying local agreement to the MT stage. All experiments were run with batch size 1 on an NVIDIA RTX A6000 GPU to ensure reproducibility.

\subsection{Decoding}
\label{sec:inference}

All the inferences are done by setting max\_new\_tokens to $256$ and beam size with num\_beams set to $5$. The same decoding settings were also followed when evaluating other models. All the inferences ran on 1 NVIDIA RTX A6000 GPU.

\begin{table*}[!t]
\resizebox{2\columnwidth}{!}{
\centering
\small
\setlength{\tabcolsep}{6pt}
\begin{tabular}{@{}c|ccccccc@{}}
\toprule
\textbf{Category} & \textbf{Model Name} & \textbf{Image} & \textbf{XCOMET-XL} & \textbf{Minor} & \textbf{Major} & \textbf{Critical} & \textbf{Total} \\ \midrule
\multirow{1}{*}{\textbf{\begin{tabular}[c]{@{}c@{}}Omni FT (Direct ST)\\\end{tabular}}} 
& LoRA FT          & \xmark      & 85.90    & \textbf{2047}  & 166   & 173 & 2386 \\ \midrule
\multirow{4}{*}{\textbf{\begin{tabular}[c]{@{}c@{}}Cascade Systems (ASR → MT)\\ Omni + SeedX\end{tabular}}} 
& No FT            & \xmark         & 89.66             & 2172  & 132   & 68    & 2372  \\
& No FT            & \checkmark     & 89.49             & 2115 & 133   & 76    & 2324  \\
& LoRA FT          & \xmark         & 89.75             & 2209  & 128   & 67    & 2404  \\
& LoRA FT          & \checkmark     & \textbf{90.18}    & 2135  & 118   & \textbf{56} & \textbf{2309} \\ \midrule
\multirow{5}{*}{\textbf{\begin{tabular}[c]{@{}c@{}}End-to-End (Direct ST)\\ OmniFusion\end{tabular}}}
& Mid Fusion            & \xmark         & 88.44             & 2126  & 157   & 74    & 2357  \\
& Mid Fusion            & \checkmark     & 90.03             & 2256  & 104   & 68    & 2428  \\
& Gated Fusion          & \xmark         & 88.09             & 2133  & 181   & 87    & 2401  \\
& Gated Fusion          & \checkmark     & 89.90             & 2217  & 106   & 60    & 2383  \\
& Gated Fusion + Self-Cascade & \checkmark & 90.05       & 2191  & \textbf{94}  & 62    & 2347  \\ \bottomrule
\end{tabular}
}
\caption{Offline ST results en$\rightarrow$de on the MCIF test. \checkmark\ indicates models using image input. Bold indicates best XCOMET-XL and lowest error counts.}
\label{tab:en_de_xcomet}
\end{table*}

\begin{table*}[!t]
\resizebox{2\columnwidth}{!}{
\centering
\small
\setlength{\tabcolsep}{6pt}
\begin{tabular}{@{}c|ccccccc@{}}
\toprule
\textbf{Category} & \textbf{Model Name} & \textbf{Image} & \textbf{XCOMET-XL} & \textbf{Minor} & \textbf{Major} & \textbf{Critical} & \textbf{Total} \\ \midrule
\multirow{1}{*}{\textbf{\begin{tabular}[c]{@{}c@{}}Omni FT (Direct ST)\\\end{tabular}}} 
& LoRA FT          & \xmark       & 80.88    & \textbf{727}  & 1018   & 54 & 1799 \\ \midrule
\multirow{4}{*}{\textbf{\begin{tabular}[c]{@{}c@{}}Cascade Systems (ASR → MT)\\ Omni + SeedX\end{tabular}}} 
& No FT            & \xmark         & 86.03             & 850   & 932   & 51    & 1833  \\
& No FT            & \checkmark     & 86.11             & 862   & 891   & 67    & 1820  \\
& LoRA FT          & \xmark         & 86.76             & 884   & 855   & 54    & 1793  \\
& LoRA FT          & \checkmark     & 86.85             & 868   & 855   & 55    & 1778  \\ \midrule
\multirow{5}{*}{\textbf{\begin{tabular}[c]{@{}c@{}}End-to-End (Direct ST)\\ OmniFusion\end{tabular}}}
& Mid Fusion            & \xmark         & 82.80             & 754           & 1065          & 93            & 1912  \\
& Mid Fusion            & \checkmark     & 85.75             & 899           & 903           & 64            & 1866  \\
& Gated Fusion          & \xmark         & 84.06             & 745  & 1085          & 56            & 1886  \\
& Gated Fusion          & \checkmark     & 86.79             & 914           & 847           & 48            & 1809  \\
& Gated Fusion + Self-Cascade & \checkmark & \textbf{87.20} & 879           & \textbf{838}  & \textbf{44}   & \textbf{1761} \\ \bottomrule
\end{tabular}
}
\caption{Offline ST results en$\rightarrow$it on the MCIF test. \checkmark\ indicates models using image input. Bold indicates best XCOMET-XL and lowest error counts.}
\label{tab:en_it_xcomet}
\end{table*}

\begin{table*}[!t]
\resizebox{2\columnwidth}{!}{
\centering
\small
\setlength{\tabcolsep}{6pt}
\begin{tabular}{@{}c|ccccccc@{}}
\toprule
\textbf{Category} & \textbf{Model Name} & \textbf{Image} & \textbf{XCOMET-XL} & \textbf{Minor} & \textbf{Major} & \textbf{Critical} & \textbf{Total} \\ \midrule
\multirow{1}{*}{\textbf{\begin{tabular}[c]{@{}c@{}}Omni FT (Direct ST)\\\end{tabular}}} 
& LoRA FT          & \xmark      & 80.13    & 463  & 1372   & \textbf{44} & 1879 \\ \midrule
\multirow{4}{*}{\textbf{\begin{tabular}[c]{@{}c@{}}Cascade Systems (ASR → MT)\\ Omni + SeedX\end{tabular}}} 
& No FT            & \xmark         & 81.94             & 615   & 1255  & 104   & 1974  \\
& No FT            & \checkmark     & 82.23             & 597   & 1254  & 81    & 1932  \\
& LoRA FT          & \xmark         & \textbf{82.74}    & 624   & 1271  & 104   & 1999  \\
& LoRA FT          & \checkmark     & \textbf{82.74}    & 604   & 1326  & \textbf{56} & 1986  \\ \midrule
\multirow{5}{*}{\textbf{\begin{tabular}[c]{@{}c@{}}End-to-End (Direct ST)\\ OmniFusion\end{tabular}}}
& Mid Fusion            & \xmark         & 79.47             & \textbf{462} & 1343  & 50    & 1855  \\
& Mid Fusion            & \checkmark     & 82.15             & 614   & 1240  & 53    & 1907  \\
& Gated Fusion          & \xmark         & 79.78             & 480   & 1298  & \textbf{45} & \textbf{1823} \\
& Gated Fusion          & \checkmark     & 82.03             & 636   & 1255  & 63    & 1954  \\
& Gated Fusion + Self-Cascade & \checkmark & 82.47       & 623   & \textbf{1225} & 60    & 1908  \\ \bottomrule
\end{tabular}
}
\caption{Offline ST results en$\rightarrow$zh on the MCIF test. \checkmark\ indicates models using image input. Bold indicates best XCOMET-XL and lowest error counts.}
\label{tab:en_zh_xcomet}
\end{table*}

\clearpage

\begin{table*}[h!]
\centering
\begin{minipage}{0.48\textwidth}
\centering
\begin{tabular}{@{}ll@{}}
\toprule
\textbf{Language} & \textbf{Code} \\ \midrule
Arabic            & \textit{ar} \\
Chinese           & \textit{zh} \\
Croatian          & \textit{hr} \\
Czech             & \textit{cs} \\
Danish            & \textit{da} \\
Dutch             & \textit{nl} \\
Finnish           & \textit{fi} \\
French            & \textit{fr} \\
German            & \textit{de} \\
Hungarian         & \textit{hu} \\
Indonesian        & \textit{id} \\
Italian           & \textit{it} \\
Japanese          & \textit{ja} \\
Korean            & \textit{ko} \\
Malay             & \textit{ms} \\
Norwegian         & \textit{no} \\
Norwegian Bokmål  & \textit{nb} \\
Polish            & \textit{pl} \\
Portuguese        & \textit{pt} \\
Romanian          & \textit{ro} \\
Russian           & \textit{ru} \\
Spanish           & \textit{es} \\
Turkish           & \textit{tr} \\
Urdu              & \textit{ur} \\
Vietnamese        & \textit{vi} \\ \bottomrule
\end{tabular}
\caption{Language codes denoted in our paper following \citet{Cheng2025-id}.}
\label{tab:lang-codes}
\end{minipage}
\hfill
\begin{minipage}{0.48\textwidth}
\centering
\resizebox{\columnwidth}{!}{
\begin{tabular}{@{}ll@{}}
\toprule
\textbf{Hyperparameter} & \textbf{Value} \\ \midrule
Gradient checkpointing & True \\
Gradient checkpointing kwargs & \texttt{\{'use\_reentrant': False\}} \\
Save steps & 1000 \\
Evaluation steps & 1000 \\
Save strategy & steps \\
Evaluation strategy & steps \\
Logging steps & 5 \\
Logging strategy & steps \\
Learning rate & 1e-4 \\
Train batch size (per device) & 4 \\
Eval batch size (per device) & 4 \\
Gradient accumulation steps & 2 \\
Weight decay & 0.01 \\
Save total limit & 3 \\
Max steps & 20000 \\
bf16 & True \\
peft type & LORA \\
qalora\_group\_size & 16 \\
r & 16 \\
target\_modules & q,k,v,up,down,gate projection \\
Push to hub & False \\
Metric for best model & loss \\
Remove unused columns & False \\
Label names & \texttt{["labels"]} \\
Dataloader num workers & 1 \\
DDP find unused parameters & True \\ \bottomrule
\end{tabular}
}
\caption{Huggingface trainer hyperparameters used for the model on 4 Nvidia A100 48GB GPUs with DDP training mode. Unspecified hyperparameters are set to default}
\label{tab:training-hyperparameters}
\end{minipage}
\end{table*}

\clearpage

\begin{table*}[!ht]
\centering
\small
\setlength{\tabcolsep}{4pt} 
\begin{tabular}{@{}c|>{\centering\arraybackslash}m{3.5cm}|>{\centering\arraybackslash}m{3.5cm}|>{\centering\arraybackslash}m{3.5cm}@{}}
\toprule
\textbf{Image} & \textbf{Cascaded FT Output} & \textbf{OmniFusion Output} & \textbf{Reference} \\ \midrule
\includegraphics[width=5cm]{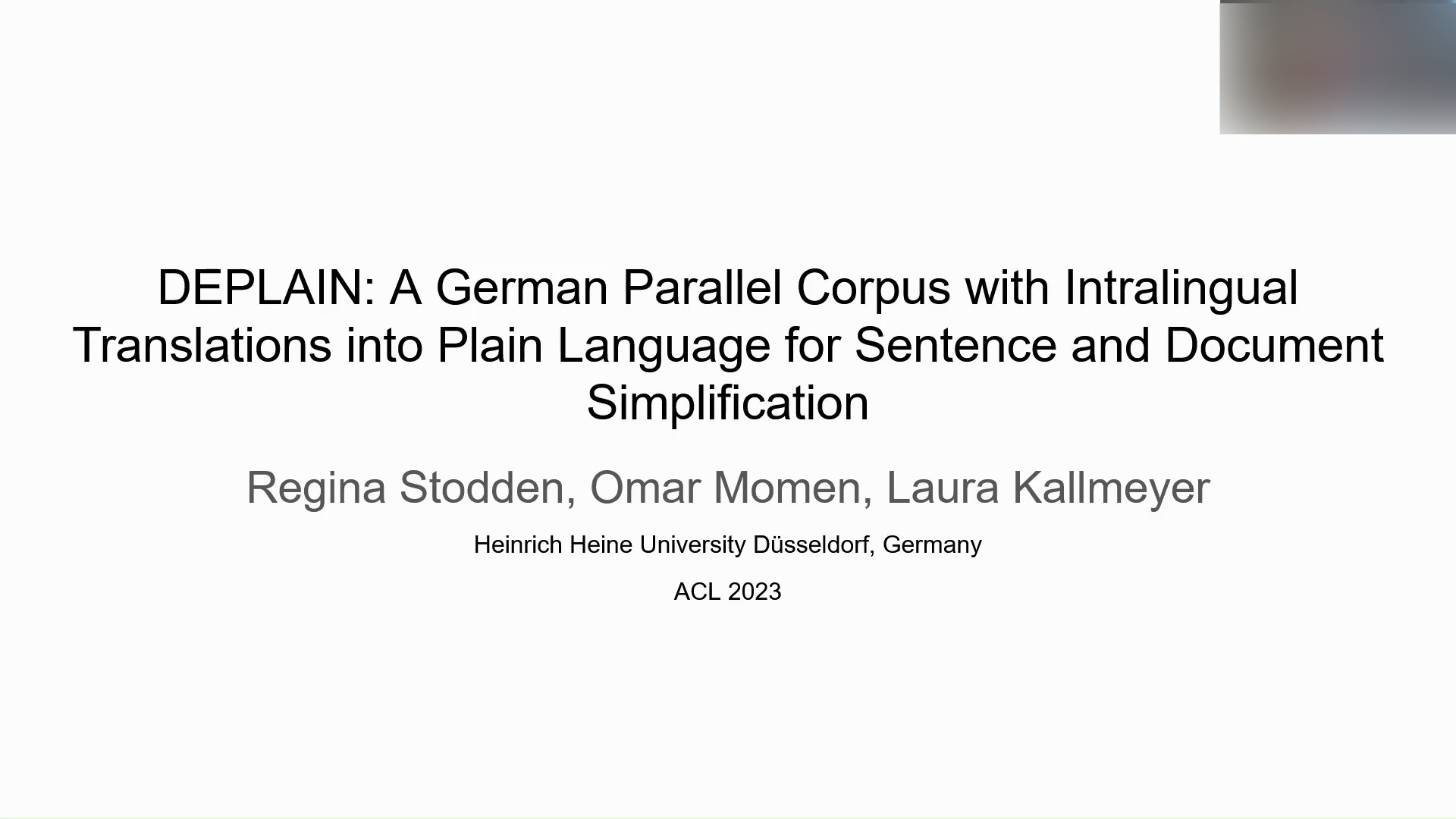} &
Hallo, willkommen zu unserer Präsentation von \textbf{Deplain}, einem neuen Korpus für die Identifizierung deutscher Texte auf Dokumenten- und Satzebene. &
Hallo, willkommen zu unserer Präsentation von \textbf{DEPLAIN}, einem neuen Korpus für die deutsche Texterkennung auf Dokumentebene und auf Satzebene. &
Herzlich willkommen zu unserer Präsentation von \textbf{DEPLAIN}, einem neuen Korpus für die deutsche Texterkennung auf Dokument- und Satzebene. \\ 
\bottomrule
\end{tabular}
\caption{Example illustrating where OmniFusion is better than cascaded variant. Example taken from MCIF segment 1.}
\label{tab:deplain_example}
\end{table*}

\begin{table*}[!ht]
\centering
\small
\setlength{\tabcolsep}{4pt} 
\begin{tabular}{@{}c|>{\centering\arraybackslash}m{3.5cm}|>{\centering\arraybackslash}m{3.5cm}|>{\centering\arraybackslash}m{3.5cm}@{}}
\toprule
\textbf{Image} & \textbf{SeedX Output} & \textbf{OmniFusion Output} & \textbf{Reference} \\ \midrule
\includegraphics[width=1cm]{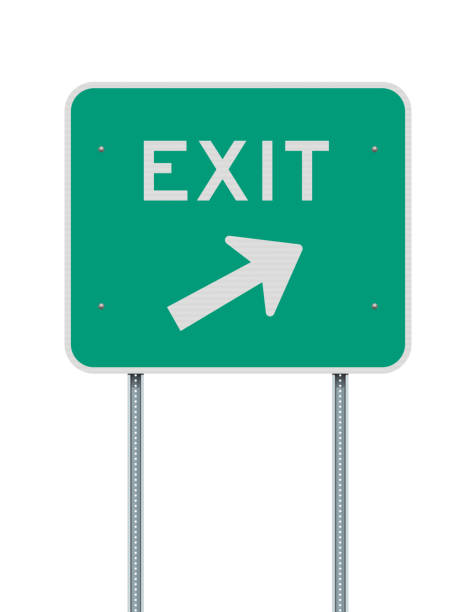} &
AUSGANG &
AUSFAHRT &
AUSFAHRT \\ 
\midrule
\includegraphics[width=1cm]{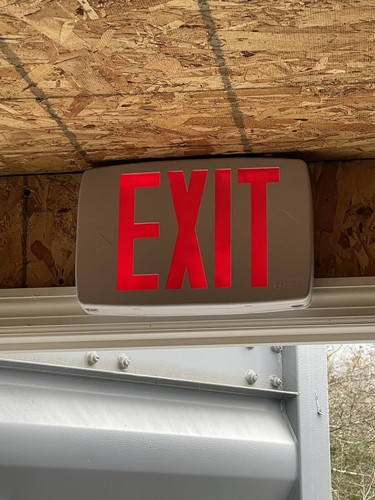} &
AUSGANG &
AUSGANG &
AUSGANG \\ 
\bottomrule
\end{tabular}
\caption{Example illustrating OmniFusion disambiguates source sentence with image context. The word 'EXIT' when translated to German can be 'AUSFAHRT' (as vehicle exit) or 'AUSGANG' (as person exit) and needs the image for disambiguation.}
\label{tab:commute_example}
\end{table*}

\end{document}